\documentclass{article}

 \usepackage[nonatbib,preprint]{neurips_2026}

\usepackage[utf8]{inputenc} 
\usepackage[T1]{fontenc}    
\usepackage{hyperref}       
\usepackage{url}            
\usepackage{booktabs}       
\usepackage{amsfonts}       
\usepackage{nicefrac}       
\usepackage[numbers]{natbib}
\usepackage{graphicx}
\usepackage{amsmath}
\usepackage{microtype}      
\usepackage{float}
\usepackage{bm}
\usepackage{multirow}
\usepackage{multicol}
\usepackage{cleveref}
\usepackage{tcolorbox}
\usepackage{amssymb}
\usepackage{pifont}
\usepackage{subcaption}

\usepackage[table,dvipsnames]{xcolor} 
\usepackage{colortbl}
\definecolor{ourblue}{RGB}{219, 234, 254}  
\definecolor{loragray}{RGB}{243, 244, 246} 

\newcommand{\tok}[2]{\begingroup
  \setlength{\fboxsep}{1pt}%
  \colorbox{#1!30}{#2}%
\endgroup}

\newcommand{\model}{BoostLLM}

\title{BoostLLM: Boosting-inspired LLM Fine-tuning for Few-shot Tabular Classification}

\author{%
  Yi-Siang Wang\thanks{Corresponding author} \quad
  Kuan-Yu Chen\thanks{Project Leader} \quad
  Yu-Chen Den \quad Darby Tien-Hao Chang \\ 
  SinoPac Holdings \\
  \texttt{\{richardwang,lavamore,abnerden,darby\}@sinopac.com}
}

\begin{document}

\maketitle

\begin{abstract}
Large language models (LLMs) have recently been adapted to tabular prediction by serializing structured features into natural language, but their performance in low-data regimes remains limited compared to gradient boosted decision trees (GBDTs). 
In this work, we revisit the boosting paradigm, traditionally associated with tree ensembles, and ask whether it can be applied as a general training principle for LLM fine-tuning. 
We propose \model{}, a single framework that transforms parameter-efficient fine-tuning (PEFT) into a multi-round residual optimization process by training sequential PEFT adapters as weak learners, where each learner is shown to provide steady performance corrections. 
To incorporate tabular inductive bias, \model{} integrates decision-tree paths as a second input view, which acts as a structured teacher in early training steps before the model shifts toward feature-driven representations.
Empirically, \model{} achieves consistent improvements over standard fine-tuning across multiple LLM backbones and datasets, matching or surpassing XGBoost across a wide range of shot counts and outperforming
state-of-the-art
GPT-4o-based methods with a 4B model. 
We further show that the framework scales: pairing with stronger tree models and extended boosting horizons yields additional gains under appropriate stabilization. 
These results suggest that boosting can serve as a general training principle for LLM fine-tuning, particularly in low-data regimes for structured data.
\end{abstract}

\section{Introduction}
\label{sec:introduction}

Tabular data serves as a primary engine for real-world decision making across domains such as finance, healthcare, and e-commerce~\citep{ma2025tabdpt,tschalzev2024datacentric}. 
Despite its ubiquity, many impactful applications operate in strictly data-constrained environments where labeled examples are scarce, expensive to obtain, or subject to rapid distribution shifts like detecting fraud, diagnosing rare diseases from limited patient records, or forecasting demand for emerging product categories. 
In such scenarios, accurate predictions must be learned from only a small number of labeled instances, making data-efficient learning a central challenge in tabular prediction~\citep{tschalzev2024datacentric,nam2023stunt,liu2024d2r2,tabred2025}.

Recent works have explored Large Language Models (LLMs) for tabular prediction. Pretrained on large-scale corpora, LLMs encode rich semantic knowledge and reasoning capabilities that can complement limited supervision in downstream tasks. 
A prominent line of work leverages LLMs for direct label prediction and feature engineering without parameter updates, relying on prompt-based inference and In-Context Learning (ICL)~\citep{one-llm-is-not-enough-2023,nam2024optimized,han2024large,ye2025llm}---an approach made increasingly viable as frontier models grow in capability.
However, fine-tuning provides a more direct adaptation pathway and has been shown to substantially outperform ICL in few-shot settings~\citep{tabllm-2022,liu-2022-tfew,dinh2022lift}.
In practice, fine-tuning is commonly performed via Parameter-Efficient Fine-Tuning (PEFT) techniques~\citep{liu-2022-tfew,hu2022lora,liao20243}, which introduce lightweight trainable modules while keeping most pretrained parameters fixed.
Despite their efficiency, such adaptation can be unstable and prone to overfitting when supervision is scarce~\citep{yaras2024compressible,gao2025optimization,lai2025joint}, highlighting the need for more data-efficient and robust fine-tuning strategies.

Interestingly, the tabular learning community has long relied on a different paradigm to address similar challenges: gradient boosting. 
Systems such as XGBoost~\citep{chen2016xgboost}, LightGBM~\citep{ke2017lightgbm}, and CatBoost~\citep{prokhorenkova2018catboost} consistently achieve strong performance across a wide range of tabular tasks and are known to be particularly robust in data-limited settings~\citep{gorishniy2021revisiting,grinsztajn2022why,ye2024closer,erickson2025tabarena,tabred2025}. 
Boosting constructs models in a stage-wise manner, where learners are added sequentially to correct the residual errors of previous ones. 
This additive training strategy encourages models to focus on informative residual errors while controlling overfitting, resulting in strong sample efficiency.
Although typically associated with tree-based models, boosting is a general paradigm that can be extended to other function classes~\citep{wang2015functional,patron2024gradient,xu2007adarank}.
This observation raises a natural question: 
\emph{Can boosting serve as a general training principle for LLM fine-tuning, particularly in low-data tabular settings?}

In this paper, we propose \textbf{\model}, a framework that instantiates boosting as a training principle for LLM fine-tuning in tabular prediction.
\model{} trains a sequence of weak learners implemented as PEFT-adapted LLMs in a stage-wise manner, where each learner focuses on correcting the residual errors of the ensemble prediction formed by previous ones. 
Each learner is trained for only a small number of epochs, limiting overfitting to the scarce supervision while encouraging each stage to capture complementary residual patterns. 
This controlled optimization preserves a training cost comparable to standard PEFT fine-tuning while improving performance in low-data regimes.
To further stabilize and enrich the boosting process, we introduce a decision-path integration strategy that incorporates structural priors from a trained gradient boosting tree.
Rather than simply concatenating decision-path descriptions with the serialized tabular input, we treat tree-derived signals as an auxiliary view of the data.
At each round, the LLM processes both a feature-only view and a path-informed view of the same sample, and their hidden states are fused before computing the class logit.
This paired-view fusion integrates structural guidance from tree-based models while allowing the LLM to progressively reduce its reliance on the tree signal as it learns.
Our contributions are as follows:
\begin{itemize}
\item We view boosting as a training principle for LLM fine-tuning in few-shot tabular prediction, and propose \model{} to realize this idea. 
\model{} trains sequential PEFT adapters as weak learners that correct residual errors stage-wise, and introduces a paired-view ensemble that fuses feature-only and path-informed hidden states to integrate decision-path knowledge at each boosting round.

\item We conduct extensive experiments on nine tabular classification benchmarks with four LLM backbones. \model{} consistently outperforms TabLLM, matches or surpasses XGBoost across a wide range of shot counts, and exceeds state-of-the-art LLM-based methods relying on GPT-4o while using only a 4B open-source model, demonstrating that boosting-based fine-tuning is both sample-efficient and architecture-agnostic.

\item Through ablation and analyses, we confirm that \model{}'s gains stem from the boosting mechanism rather than parameter count, and that the method is robust to hyperparameter choices. Behavioral analysis shows that boosting provides consistent corrections across rounds, and that the path view serves as an early-stage teacher whose influence diminishes as the LLM matures. We also show that the framework scales with stronger tree models, yielding additional gains.

\end{itemize}
\section{Preliminary}

\paragraph{Task Formulation.}
Let $\mathcal{D}=\{(\mathbf{x}_i, y_i)\}_{i=1}^n$ denote a tabular classification dataset with $n$ samples, where $\mathbf{x}_i = [x_{i1},\ldots,x_{iK}]^\top \in \mathbb{R}^K$ is a feature vector with $K$ attributes,
and $y_i \in \mathcal{C} = \{1,\dots,C\}$ is the class label among $C$ classes. The goal is to learn a model that accurately predicts the label $y_i$ given input $\mathbf{x}_i$.

\paragraph{Gradient Boosting.}
Gradient boosting constructs additive predictors in a stage-wise manner:
\begin{equation}
    F_R(\mathbf{x}_i) = \sum_{r=1}^{R} \eta\, f_r(\mathbf{x}_i),
    \label{eq:gb}
\end{equation}
where $f_r$ is the weak learner trained at round $r$ and $\eta$ is the boosting learning rate.
Each $f_r$ is fitted to the \emph{residual} between the target and the current ensemble prediction $F_{r-1}(\mathbf{x}_i)$, such that successive rounds iteratively correct the errors of previous ones.

When instantiated with decision trees, each $f_r$ produces predictions via a root-to-leaf path, which can be interpreted as a conjunction of feature conditions (e.g., $(\texttt{age} > 30) \wedge (\texttt{income} \leq 50\mathrm{K})$).
This stage-wise, error-correcting inductive bias has proven highly effective for tabular data~\citep{chen2016xgboost, ke2017lightgbm, prokhorenkova2018catboost}. However, it has not been systematically incorporated into LLM-based tabular classifiers, which \model{} aims to address.

\paragraph{LLM-based Tabular Classification.}
To enable Large Language Models (LLMs) to process tabular inputs, each sample $\mathbf{x}_i$ is converted into a natural-language sequence.
Following TabLLM~\citep{tabllm-2022}, we serialize $\mathbf{x}_i$ using a fixed template $\mathcal{T}$, referred to as the \emph{feature-only prompt}:
\begin{equation}
    \mathcal{T}(\mathbf{x}_i)
    = \underbrace{\text{The } e_{1} \text{ is } x_{i1}.\ \cdots\
                   \text{The } e_{K} \text{ is } x_{iK}.}_{\text{serialized features}}
    \;\; \underbrace{\texttt{<Task Description>}}_{\text{prediction objective}}
    \;\; \text{Answer:},
    \label{eq:tab-template}
\end{equation}
where $e_k$ denotes the column name of the $k$-th feature, and \texttt{<Task Description>} is a dataset-specific natural-language prompt that specifies the prediction objective---for example, \textit{``Does this client subscribe to a term deposit? Answer Yes or No.''}

Following T-Few~\citep{liu-2022-tfew}, we score a set of verbalized class labels to obtain class predictions. Concretely, let $\mathbf{s}_c = (s_{c,1}, \dots, s_{c,L_c})$ denote the token sequence corresponding to the verbalized label for class $c \in \mathcal{C}$, where $L_c$ is its token length. For example, for class $c = \texttt{positive}$, $\mathbf{s}_c$ is the tokenized sequence under the model's tokenizer. Given the serialized input, the choice logit for class $c$ is computed as the length-normalized log-likelihood:
\begin{equation}
    z(c \mid \mathbf{x}_i)
    = \frac{1}{L_c} \sum_{t=1}^{L_c}
      \log p_\theta\!\left(s_{c,t} \mid \mathcal{T}(\mathbf{x}_i),\,
      \mathbf{s}_{c,<t}\right),
    \label{eq:logit}
\end{equation}
where $p_\theta$ denotes the token probability output of the LLM parameterized by $\theta$, and $\mathbf{s}_{c,<t} = (s_{c,1}, \dots, s_{c,t-1})$ is the prefix of the verbalized label. The predicted class and class probabilities are derived directly from the logits $z(c \mid \mathbf{x}_i)$ across $c \in \mathcal{C}$.

In this paper, we retain the prediction formulation in \Cref{eq:logit}. The key contribution of \model{} lies in \emph{how the per-round logits are produced and accumulated}, following the stage-wise residual fitting paradigm in \Cref{eq:gb}. The full procedure is described in \Cref{sec:method}.

\begin{figure}[t]
  \centering
  \includegraphics[width=1.0\linewidth]{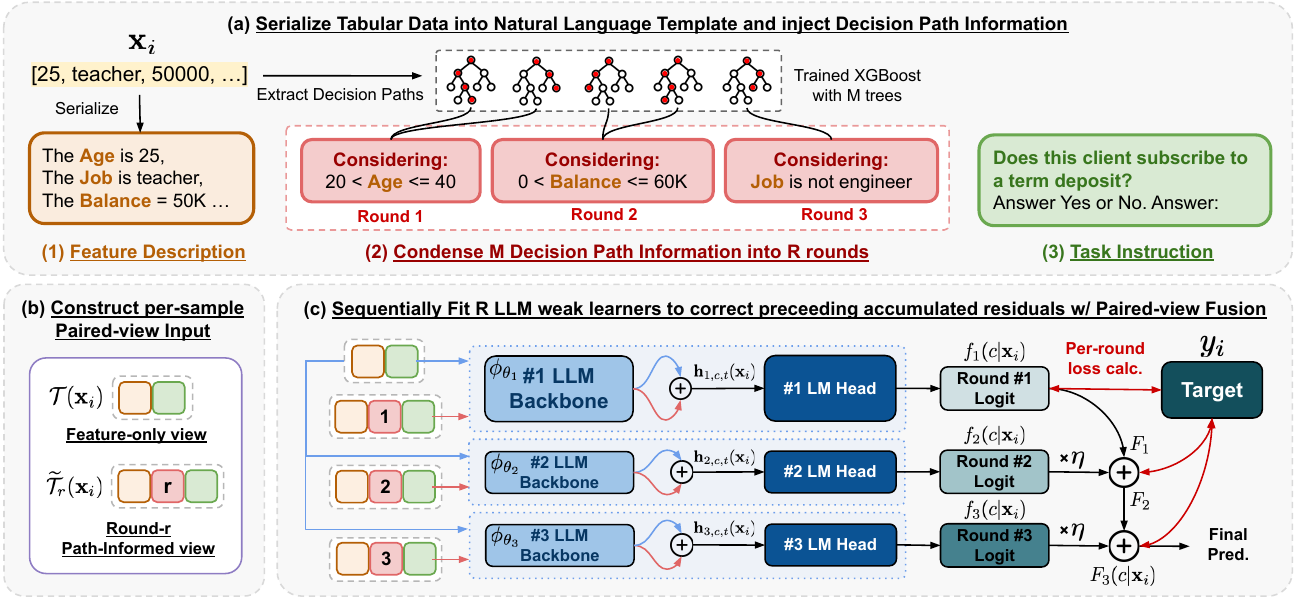}
  \caption{Overview of \model{}.
\textbf{(a)} Tabular features $\mathbf{x}_i$ are serialized into a 
natural language template. An XGBoost model's $M$ decision paths are 
condensed into $R$ round-specific descriptions via constraint intersection 
(\Cref{eq:condensed-path}).
\textbf{(b)} Each sample is presented as a feature-only and a path-informed 
prompt (\Cref{eq:tab-template,eq:tab-path-template}).
\textbf{(c)} $R$ LLM weak learners with round-specific parameters $\theta_r$ are 
trained sequentially. Both views are forwarded through backbone 
$\phi_{\theta_r}$ and their hidden states are averaged to produce per-round 
logit $f_r$ (\Cref{eq:paired-hidden,eq:round-logit}); each learner corrects 
the residual of the preceding accumulated logit $F_{r-1}$, which accumulates 
stagewise into $F_r$ (\Cref{eq:boosting}).}
  \label{fig:BoostLLM}
\end{figure}

\section{Methodology}
\label{sec:method}
\model{} instantiates the gradient boosting framework of \Cref{eq:gb} in 
logit space, using LLMs as weak learners (see \Cref{fig:BoostLLM}).
\Cref{sec:method-paths} extracts and compresses XGBoost decision paths into 
round-specific natural-language descriptions; 
\Cref{sec:method-boosting} describes how each round trains a 
parameter-efficient adapter on paired views of each sample and accumulates 
the resulting logit corrections stagewise; 
\Cref{sec:training-objective} presents the training objective and inference 
procedure.

\subsection{Decision-Path Extraction and Compression}
\label{sec:method-paths}

We first train an XGBoost model with $M$ trees on the same training split used for \model{}. For each sample $\mathbf{x}_i$, we extract the root-to-leaf decision path from each of the $M$ trees as a set of feature constraints, yielding $\{\pi_m(\mathbf{x}_i)\}_{m=1}^{M}$.

Since fine-tuning an LLM for each of the $M$ trees is computationally prohibitive, we compress them into $R$ round-specific descriptions where $R \ll M$. Concretely, we partition the $M$ paths in their original boosting order into $R$ consecutive groups of approximately equal size, and let $\mathcal{G}_r$ denote the indices of trees assigned to round $r$. 
For each round $r$, we merge the paths in $\mathcal{G}_r$ by intersecting their constraints, yielding a condensed path $\pi^*_r(\mathbf{x}_i)$:
\begin{equation}
    \pi^*_r(\mathbf{x}_i) = \bigwedge_{m \in \mathcal{G}_r} \pi_m(\mathbf{x}_i).
    \label{eq:condensed-path}
\end{equation}
A worked example is provided in \Cref{sec:appendix-path-example}.
Applying this procedure across all $R$ rounds yields $\{\pi^*_r(\mathbf{x}_i)\}_{r=1}^{R}$, providing one condensed decision path per round. This compression keeps prompt length manageable while preserving key feature constraints learned by the trees.

\subsection{Stage-wise Boosting of LLM Weak Learners}
\label{sec:method-boosting}

\model{} sequentially trains $R$ weak learners $\{f_r\}_{r=1}^{R}$, each sharing a pretrained LLM backbone but equipped with a round-specific RoAd adapter~\citep{liao20243}; we denote the combined parameters at round $r$ by $\theta_r$.
At each round $r$, all previous models $\{f_{r'}\}_{r' < r}$ are frozen , while only $f_r$ is optimized to correct the residual of the accumulated logits $F_{r-1}(\mathbf{x}_i)$, following \Cref{eq:gb}.

\paragraph{Paired-view inputs.}
At each round $r$, every sample is presented under two views. The first is the feature-only prompt $\mathcal{T}(\mathbf{x}_i)$ from~\Cref{eq:tab-template}, and the second is the \emph{path-informed prompt} $\widetilde{\mathcal{T}}_r(\mathbf{x}_i)$, constructed by converting the condensed path $\pi^*_r(\mathbf{x}_i)$ into a natural language string and inserting it between the serialized features and the task description:
\begin{equation}
    \widetilde{\mathcal{T}}_r(\mathbf{x}_i)
    = \text{The } h_1 \text{ is } x_{i1}.\ \cdots\
      \text{The } h_F \text{ is } x_{iF}.\
      \text{Considering } \pi^*_r(\mathbf{x}_i).\
      \texttt{<Task Description>}\ \text{Answer:}
    \label{eq:tab-path-template}
\end{equation}

\paragraph{Class logit computation.}
For each view and each class $c$, we obtain hidden states at each class token position $t \in \{1, \dots, L_c\}$ from the round-$r$ LLM backbone $\phi_{\theta_r}$.
We then compute a fused hidden state by averaging:
\begin{equation}
    \mathbf{h}_{r,c,t}(\mathbf{x}_i)
    = \frac{1}{2}\Bigl(
        \phi_{\theta_r}\bigl(\mathcal{T}(\mathbf{x}_i) \| \mathbf{s}_{c,<t}\bigr) +
        \phi_{\theta_r}\bigl(\widetilde{\mathcal{T}}_r(\mathbf{x}_i) \| \mathbf{s}_{c,<t}\bigr)
      \Bigr),
    \label{eq:paired-hidden}
\end{equation}
where $\|$ denotes concatenation. The round-$r$ class logit is then computed as the length-normalized log-likelihood over the fused hidden states following~\Cref{eq:logit}:
\begin{equation}
    f_r(c \mid \mathbf{x}_i)
    = \frac{1}{L_c} \sum_{t=1}^{L_c}
      \log p_{\theta_r}\!\left(s_{c,t} \mid \mathbf{h}_{r,c,t}(\mathbf{x}_i)\right).
    \label{eq:round-logit}
\end{equation}
\paragraph{Boosting accumulation.}
With the per-round logit $f_r(c \mid \mathbf{x}_i)$ computed, we update the accumulated logit $F_r(c \mid \mathbf{x}_i)$ following \Cref{eq:gb} with boosting learning rate $\eta$:
\begin{equation}
    F_r(c \mid \mathbf{x}_i) =
    \begin{cases}
        0 & r = 0,\\
        F_{r-1}(c \mid \mathbf{x}_i) + \eta\, f_r(c \mid \mathbf{x}_i) & r \geq 1.
    \end{cases}
    \label{eq:boosting}
\end{equation}

\subsection{Training Objective and Inference}
\label{sec:training-objective}

At round $r$, learner $f_r$ is trained with the same three-term objective as T-Few~\citep{liu-2022-tfew}:
$\mathcal{L} = \mathcal{L}_{\mathrm{LM}} + \mathcal{L}_{\mathrm{UL}} + \mathcal{L}_{\mathrm{CLS}}$.
The key difference is that $\mathcal{L}_{\mathrm{CLS}}$ is applied to the \emph{accumulated} logits $F_r(\cdot \mid \mathbf{x}_i)$ rather than the logits of a standalone model, so each learner explicitly corrects the residuals of previous rounds.
For binary classification, $\mathcal{L}_{\mathrm{CLS}}$ is binary cross-entropy on $F_r(y^{+} \mid \mathbf{x}_i)$, where $y^{+} \in \mathcal{C}$ denotes the positive class; for multiclass classification, it is cross-entropy over $F_r(\cdot \mid \mathbf{x}_i) \in \mathbb{R}^{|\mathcal{C}|}$.
At test time, inference follows the procedure of \Cref{eq:boosting}. The final prediction is $\sigma(F_R(y^{+} \mid \mathbf{x}_i))$ for binary classification and $\mathrm{softmax}(F_R(\cdot \mid \mathbf{x}_i))$ for multiclass classification.
\section{Experiments}
\label{sec:experiments}

\subsection{Experimental Setup}
\label{sec:exp-setup}

\paragraph{Datasets.}
For most experiments, we follow TabLLM~\citep{tabllm-2022} and evaluate on nine datasets: Bank~\citep{moro2014bank}, Blood~\citep{yeh2008blood}, California~\citep{grinsztajn2022why}, Car~\citep{bohanec1988knowledge}, Credit-g~\citep{hofmann1994statlog}, Diabetes~\citep{smith1988using}, Heart~\citep{janosi1988heart}, Income~\citep{kohavi1996scaling}, and Jungle~\citep{van2014endgame}.
For comparison 
in \Cref{sec:exp-sota},
we follow DeLTa~\citep{ye2025llm} and evaluate on Blood, Car, Credit-g, Diabetes, Heart, and Jannis~\citep{grinsztajn2022why}.
Dataset statistics are in \Cref{sec:appendix-datasets}.

\paragraph{Baselines.}
We compare \model{} against:
(1)~\textbf{Gradient boosting trees}: XGBoost~\citep{chen2016xgboost}, with hyperparameters tuned via Optuna~\citep{akiba2019optuna} on validation performance (search space in \Cref{sec:appendix-xgboost}); 
(2)~\textbf{LLM-based methods}: TabLLM~\citep{tabllm-2022} is re-evaluated under our protocol using the same hyperparameter settings
as \model{}, except in \Cref{sec:exp-sota}, where results for TabLLM along with LIFT~\citep{dinh2022lift}, LIFT-ICL~\citep{dinh2022lift}, TP-BERTa~\citep{yan2024making}, GTL~\citep{wen2024supervised}, P2T~\citep{nam2024tabular}, FeatLLM~\citep{han2024large}, and DeLTa are taken directly from \citet{ye2025llm}.

\paragraph{Backbone LLMs.}
\model{} is applied to two LLM families---T5Gemma2 (encoder–decoder)~\citep{zhang2025t5gemma} and Qwen3 (decoder-only)~\citep{yang2025qwen3}---each at 4B and 8B parameters.
Following TabLLM, all models are trained with a fixed set of hyperparameters without per-dataset tuning.
Full hyperparameter settings and training details are provided in \Cref{sec:appendix-hparams}.

\paragraph{Evaluation Protocol.}
All models are evaluated using 5-fold stratified cross-validation, with shot counts (i.e., the number of training samples) varying by experiment:
comparison with state-of-the-art LLM-based methods (\Cref{sec:exp-sota}) follows DeLTa with 64, 256, and full shots;
GBDT and fine-tuning comparison (\Cref{sec:exp-main})
uses from 4 shots to the full training set;
analysis experiments (\Cref{sec:ablation-study,sec:analysis,sec:scaling}) use 128 shots. The primary metric is average precision (AP), with macro average applied for multi-class tasks.
When accuracy is reported (\Cref{sec:exp-sota}), the decision threshold is calibrated on the training set to account for logit scale variation.
Scores in the main text are averaged across datasets; per-dataset results with standard deviations are provided in the appendix.

\subsection{Comparison with State-of-the-Art LLM-based Methods}
\label{sec:exp-sota}
A natural question is whether a fine-tuned open-source model can match or surpass
the state-of-the-art LLM-based methods.
As shown in~\Cref{tab:compare-others}, 
\model{} (Qwen3-4B) achieves the highest average accuracy in all three settings, outperforming DeLTa (GPT-4o)---the current state-of-the-art training-free method---by 6.1, 6.6, and 3.0 points at 64, 256, and all shots, respectively---indicating that boosting-based fine-tuning of a 4B open-source model can surpass approaches relying on larger proprietary models. Per-dataset results are in~\Cref{sec:appendix-per-dataset}.

\subsection{Comparison with GBDT and Vanilla Fine-tuning}
\label{sec:exp-main}
Beyond comparing individual methods, we examine two questions: 
(1) whether \model{} closes the gap with GBDT methods in the moderate-to-large data regime, and 
(2) whether boosting-based fine-tuning provides consistent gains over vanilla LLM fine-tuning across architectures.
\Cref{tab:compare-llm-gbdt} reports AP across shot counts from 4 to all shots.
\model{} consistently outperforms TabLLM across all backbones and data regimes, confirming that the boosting paradigm provides reliable gains over standard fine-tuning regardless of model family.
For at least one backbone, \model{} matches or exceeds XGBoost at every shot count,
with gains consistent across both encoder--decoder and decoder-only families.
Per-dataset results are in~\Cref{sec:appendix-shot-tables}.

\begin{table}[t]
  \caption{Average test accuracy comparison with state-of-the-art LLM-based tabular classification methods. All baselines are from DeLTa~\citep{ye2025llm}. GTL, LIFT-ICL, and P2T are omitted from 256- and All-shot as these results are not reported in the corresponding papers. \textbf{Bold}: best per row; \underline{Underline}: second best per row. $^\ast$All-shots excludes Bank, Income, and Jannis due to large training set sizes.}
  \label{tab:compare-others}
  \vspace{0.5em}
  \centering
  \resizebox{\columnwidth}{!}{%
    \begin{tabular}{l|ccccccccc}
      \toprule
      \textbf{Method}
        & TabLLM & LIFT & TP-BERTa & GTL & LIFT-ICL & P2T & FeatLLM & DeLTa & \textbf{\model{}} \\
      \textbf{LLM Used}
        & T0 & GPT-3.5 & RoBERTa & LLaMA2 & GPT-4o & GPT-4o & GPT-4o & GPT-4o & Qwen3-4B \\
      \midrule
      64-shot  & 49.3 & 61.5 & 62.1 & 60.8 & 42.4 & 44.8 & 62.8 & \underline{66.0} & \textbf{72.1} \\
      256-shot & 57.5 & 61.0 & 64.9 & -    & -    & -    & 63.9 & \underline{70.0} & \textbf{76.6} \\
      $^\ast$All-shot & 78.1 & 70.3 & 77.2 & -    & -    & -    & 68.6 & \underline{81.6} & \textbf{84.6} \\
      \bottomrule
    \end{tabular}
  }
\end{table}

\begin{table*}[t]
  \caption{Average Precision (AP) across varying shot counts. 
\textbf{Bold} and \underline{underline} denote the best and second-best performance across all methods for each shot setting, respectively. \tok{gray}{Shaded cells} indicate the better method between TabLLM and \model{} under the same LLM backbone. $^\ast$All-shots excludes Income, California, and Jungle due to large training set sizes.}
  \label{tab:compare-llm-gbdt}
  \centering
  \resizebox{0.8\textwidth}{!}{%
  \begin{tabular}{ll|ccccccccc} 
    \toprule
    \multirow{2}{*}{\textbf{LLM used}} & \multirow{2}{*}{\textbf{Method}} & \multicolumn{9}{c}{\textbf{Number of Shots ($N$)}} \\
    & & \textbf{4} & \textbf{8} & \textbf{16} & \textbf{32} & \textbf{64} & \textbf{128} & \textbf{256} & \textbf{512} & \textbf{*All} \\
    \midrule
    -- & XGBoost & 57.9 & 59.9 & 65.5 & 70.4 & 75.4 & 78.5 & 81.3 & 83.0 & \underline{83.0} \\
    \midrule

    \multirow{2}{*}{T5Gemma2-2B} 
    & TabLLM 
    & 57.6 & \cellcolor{gray!35}61.2 & 64.3 & 68.8 & 71.0 & 74.2 & 76.6 & 77.3 & 76.9 \\
    & \textbf{\model{}} 
    & \cellcolor{gray!35}59.6 & 60.8 & \cellcolor{gray!35}66.3 & \cellcolor{gray!35}70.1 & \cellcolor{gray!35}75.4 & \cellcolor{gray!35}79.1 & \cellcolor{gray!35}\underline{81.7} & \cellcolor{gray!35}\textbf{83.5} & \cellcolor{gray!35}\textbf{83.3} \\
    \midrule

    \multirow{2}{*}{T5Gemma2-8B} 
    & TabLLM 
    & 59.4 & 60.4 & 64.3 & 67.8 & 71.9 & 74.6 & 77.2 & 77.6 & 77.4 \\
    & \textbf{\model{}} 
    & \cellcolor{gray!35}\underline{60.0} & \cellcolor{gray!35}61.9 & \cellcolor{gray!35}66.5 & \cellcolor{gray!35}\underline{71.0} & \cellcolor{gray!35}76.4 & \cellcolor{gray!35}79.1 & \cellcolor{gray!35}81.6 & \cellcolor{gray!35}83.3 & \cellcolor{gray!35}\underline{83.0} \\
    \midrule

    \multirow{2}{*}{Qwen3-4B} 
    & TabLLM 
    & \cellcolor{gray!35}\underline{60.0} & 62.1 & 63.2 & 67.0 & 71.7 & 74.8 & 75.7 & 78.4 & 77.6 \\
    & \textbf{\model{}} 
    & 59.9 & \cellcolor{gray!35}\underline{63.0} & \cellcolor{gray!35}\underline{68.2} & \cellcolor{gray!35}\textbf{72.8} & \cellcolor{gray!35}\textbf{77.0} & \cellcolor{gray!35}\textbf{79.7} & \cellcolor{gray!35}\textbf{82.2} & \cellcolor{gray!35}\underline{83.4} & \cellcolor{gray!35}\underline{83.0} \\
    \midrule

    \multirow{2}{*}{Qwen3-8B} 
    & TabLLM 
    & 58.5 & 61.1 & 63.3 & 67.3 & 70.1 & 73.0 & 75.2 & 77.7 & 78.0 \\
    & \textbf{\model{}} 
    & \cellcolor{gray!35}\textbf{62.4} & \cellcolor{gray!35}\textbf{64.1} & \cellcolor{gray!35}\textbf{68.5} & \cellcolor{gray!35}\textbf{72.8} & \cellcolor{gray!35}\underline{76.9} & \cellcolor{gray!35}\underline{79.2} & \cellcolor{gray!35}81.5 & \cellcolor{gray!35}82.3 & \cellcolor{gray!35}82.6 \\
    \bottomrule
  \end{tabular}
  }
\end{table*}

\begin{table}[t]
  \caption{Ablation study. Lower rows progressively add components; TabLLM (ensemble) is a parameter-matched baseline that independently trains 5 models and averages predictions. \textbf{Bold} and \underline{underline} denote the best and second-best results, respectively. It is worth noting that all \model{} variants outperform XGBoost.}
  \vspace{0.5em}
  \label{tab:ablation}
  \centering
  \resizebox{0.85\textwidth}{!}{%
  \begin{tabular}{lcccccc}
    \toprule
    \textbf{Model} & \textbf{Boosting} & \textbf{Feature-only View} & \textbf{Path-informed View} & \textbf{Params} & \textbf{Avg. AP} \\
    \midrule
    XGBoost    & --- & --- & ---        & ---        & 78.5 \\
    TabLLM & & \checkmark & & $1\times$  & 74.8 \\
    TabLLM (ensemble) &  & \checkmark &  & $5\times$  & 76.1 \\
    \midrule
    \model{} & \checkmark & \checkmark & & $5\times$  & 78.8 \\
             & \checkmark &  & \checkmark & $5\times$  & \underline{79.1} \\
             & \checkmark & \checkmark & \checkmark & $5\times$ & \textbf{79.7} \\    
    \bottomrule
  \end{tabular}
  }
\end{table}

\subsection{Ablation Study}
\label{sec:ablation-study}

We conduct two ablations to verify that \model{}'s gains stem from its design rather than confounding factors. All experiments use Qwen3-4B with 128-shot training data.

\paragraph{Component Ablation.}
We progressively transform TabLLM into \model{} by adding one component at a time.
As shown in \Cref{tab:ablation}, introducing boosting yields the largest gain, improving AP by 4.0 points (74.8 $\rightarrow$ 78.8), already surpassing XGBoost (78.5).
Concatenating decision paths provides only a modest improvement to 79.1 (+0.3).
In contrast, combining the two views via hidden-state fusion (\Cref{eq:paired-hidden}) achieves the best performance of 79.7 (+0.9 over 78.8). 

\paragraph{Parameter Efficiency.}
To isolate the effect of the boosting mechanism from increased parameter count, we compare \model{} against a parameter-matched TabLLM ensemble with five independently trained PEFT adapters.
This ensemble averages the outputs of the five models and yields only a marginal improvement over a single TabLLM (76.1 vs.\ 74.8 AP).
In contrast, introducing boosting achieves 78.8 AP, and the final \model{} further reaches 79.7 AP, suggesting that the gains arise from the boosting mechanism rather than parameter count alone.
\section{Analysis}
\label{sec:analysis}
We analyze \model{} from two angles: whether its performance is robust to hyperparameter choices, and what mechanisms govern its internal behavior.
All experiments use Qwen3-4B with 128-shot training data.

\subsection{Hyperparameter Sensitivity}
\label{sec:hyperparam-sensitivity}
\paragraph{Round--Epoch Allocation.}
We study how to allocate a fixed total budget of 30 epochs between boosting rounds and epochs per round ($R \times E = 30$).
As shown in \Cref{tab:ablation-rounds}, only the extreme settings exhibit a noticeable drop in AP: a single round (74.6, equivalent to standard fine-tuning) and 30 rounds with one epoch each (77.9).
For all other splits between 3 and 15 rounds, AP remains within a narrow range of 1.2 points (78.7--79.9), suggesting that the round--epoch allocation does not require careful per-dataset tuning.
Full per-dataset breakdowns are provided in \Cref{sec:appendix-parameter-sensitivity}.

\paragraph{Boosting Learning Rate.}
The boosting learning rate $\eta$ controls how aggressively each learner's logits are added to the accumulated logits (\Cref{eq:gb}).
By default, \model{} uses $\eta = 0.3$, following the default setting of XGBoost.
As shown in \Cref{tab:ablation-lr}, average AP varies by less than 1 point across $\eta \in [0.1, 1.0]$, indicating that \model{} is not sensitive to this hyperparameter.
Full per-dataset results are also provided in \Cref{sec:appendix-parameter-sensitivity}.

\begin{table}[t]
  \centering
  \caption{Hyperparameter sensitivity analysis.
    \textbf{(a)} Round--epoch allocation under a fixed budget of 30 epochs.
    \textbf{(b)} Sensitivity to boosting learning rate $\eta$.
    \textbf{Bold}: best; \underline{underline}: second best.}
  \label{tab:sensitivity}

  \begin{subtable}[t]{0.55\linewidth}
    \centering
    \caption{Round--epoch allocation ($R \times E = 30$).}
    \label{tab:ablation-rounds}
    \resizebox{\linewidth}{!}{%
    \begin{tabular}{lccccccccc}
      \toprule
      \textbf{\# Rounds} & 1    & 2    & 3    & \textbf{5 (default)}      & 6             & 10               & 15   & 30   \\
      \textbf{\# Epochs} & 30   & 15   & 10   & \textbf{6 (default)}       & 5             & 3                & 2    & 1    \\
      \midrule
      \textbf{Avg.\ AP}  & 74.6 & 76.9 & 78.7 & \underline{79.7} & \textbf{79.9} & \underline{79.7} & 79.5 & 77.9 \\
      \bottomrule
    \end{tabular}}
  \end{subtable}
  \hfill
  \begin{subtable}[t]{0.42\linewidth}
    \centering
    \caption{Boosting learning rate $\eta$.}
    \label{tab:ablation-lr}
    \resizebox{\linewidth}{!}{%
    \begin{tabular}{lcccccc}
      \toprule
      \textbf{$\eta$}   & 0.1  & \begin{tabular}{@{}c@{}}\textbf{0.3}\\\textbf{(default)}\end{tabular}  & 0.5  & 0.7  & 0.9  & 1.0  \\
      \midrule
      \textbf{Avg.\ AP} & 79.1 & \textbf{79.7} & 79.5 & \underline{79.6} & 79.1 & 78.8 \\
      \bottomrule
    \end{tabular}}
  \end{subtable}
\end{table}

\subsection{Behavioral Analysis}
\label{sec:behavioral-analysis}

We examine two aspects of \model{}'s internal behavior: how boosting corrections evolve across rounds, and how the relative contribution of the two input views shifts during training.

\begin{figure}[t]
  \centering
  \resizebox{0.9\linewidth}{!}{%
  \begin{subfigure}[b]{0.5\linewidth}
    \centering
    \includegraphics[width=\linewidth]{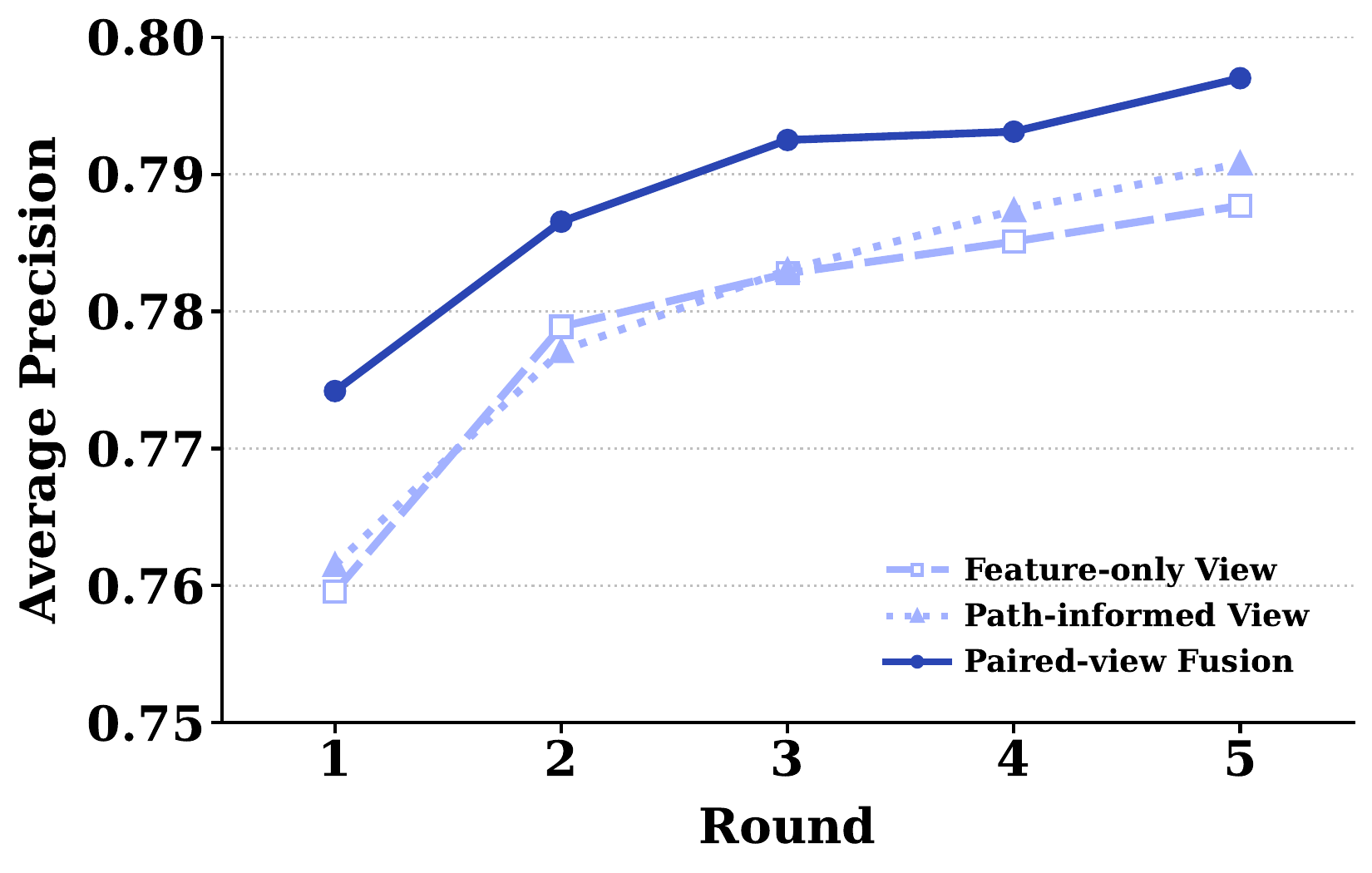}
    \phantomsubcaption
    \label{fig:boosting-dynamics}
  \end{subfigure}
  \hfill
  \begin{subfigure}[b]{0.5\linewidth}
    \centering
    \includegraphics[width=\linewidth]{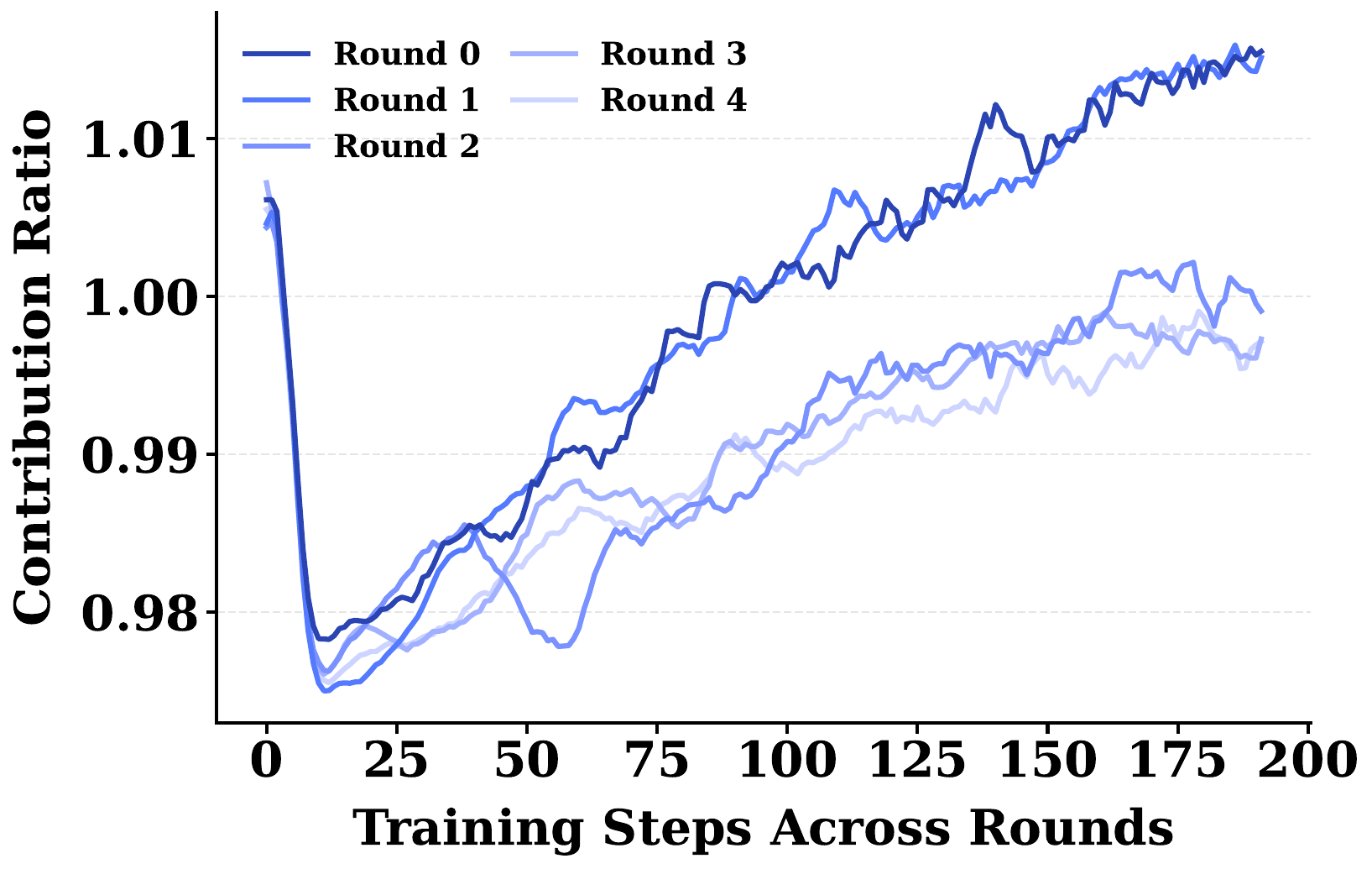}
    \phantomsubcaption
    \label{fig:magnitude-ratio}
  \end{subfigure}
  }%
  \caption{Behavioral analysis of \model{}.
  \textbf{(a)} Left: Per-round AP during boosting; dot/dashed lines denotes different views, while solid line denotes the fusion of both views.
  \textbf{(b)} Right: Per-round view contribution ratio $\rho_r$ over training steps, averaged over nine datasets; Smaller $\rho_r$ indicates stronger path-view influence, while larger $\rho_r$ indicates stronger feature-view influence.}
  \label{fig:mechanisms}
\end{figure}

\paragraph{Boosting Dynamics.}
\label{sec:boosting-dynamics}
\Cref{fig:boosting-dynamics} shows per-boosting round AP with feature-only view, path-informed view, and paired-view fusion.
AP increases steadily across rounds in all settings, confirming that boosting provides meaningful corrections at each stage.
The solid line (paired-view fusion) invariably track above others across rounds, confirming that view fusion provides an additive gain throughout training.
Per-dataset trajectories are in \Cref{sec:appendix-grid}.

\paragraph{View Contribution Dynamics.}
To quantify the relative influence of each view on the fused hidden state (\Cref{eq:paired-hidden}), we define the view contribution ratio at round $r$ as:
\begin{equation}
    \rho_r = \|\phi_{\theta_r}(\mathcal{T}(\mathbf{x})\|\mathbf{s})\| \;/\; \|\phi_{\theta_r}(\widetilde{\mathcal{T}}_r(\mathbf{x})\|\mathbf{s})\|.
\end{equation}
Since the fused state is the arithmetic mean of the two view vectors, the ratio determines each view's angular contribution to the ensemble direction: a lower $\rho_r$ indicates stronger path-view influence, while a higher $\rho_r$ indicates stronger feature-view influence.
\Cref{fig:magnitude-ratio} plots the mean ratio across nine datasets over training steps for each boosting round.
In early training of each round, $\rho$ decreases, indicating that the model initially learn knowledge from the decision-path view---the structured signal from XGBoost acts as a teacher guiding the LLM's representations.
As training continues, $\rho$ steadily increases, suggesting that, having internalized the tree-based guidance, the LLM shifts toward its own feature-driven representations with diminishing reliance on the external teacher.
Per-dataset trajectories (\Cref{sec:appendix-ratio}) confirm that, although absolute ratio magnitudes vary across datasets, most follow this dip-then-rise pattern.

\begin{table}[t]
  \caption{Scaling \model{} with XGBoost Large (\texttt{n\_estimators} up to 2000, $20\times$ the default). Logit decay $\alpha$ is applied to stabilize accumulation when boosting for 30 rounds. \textbf{Bold}: best.}
  \vspace{0.5em}
  \label{tab:scaling}
  \centering
  \resizebox{0.6\textwidth}{!}{%
  \begin{tabular}{lcc|c}
    \toprule
    \textbf{Model} & \textbf{Rounds} & \textbf{Decay $\alpha$} & \textbf{Avg.\ AP} \\
    \midrule
    XGBoost & -- & -- & 78.5 \\
    XGBoost Large & -- & -- & 79.5 \\
    \midrule
    \model{} (XGBoost) & 5 & -- & 79.7 \\
    \model{} (XGBoost) & 5 & 0.9 & 78.1 \\
    \midrule
    \model{} (XGBoost Large) & 5 & -- & 79.0 \\
    \model{} (XGBoost Large) & 30 & -- & 79.1 \\
    \model{} (XGBoost Large) & 30 & 0.9 & \textbf{80.3} \\
    \bottomrule
  \end{tabular}
  }
\end{table}

\section{Generalization to Stronger Tree Models}
\label{sec:scaling}
A natural question is whether \model{} remains effective when paired with a stronger tree model.
We investigate this using Qwen3-4B with 128-shot training data, replacing the default XGBoost (capped at 100 estimators) with XGBoost-Large (2000 estimators, $20\times$ more trees), which achieves 79.5 AP as a standalone classifier.
However, directly substituting XGBoost-Large into \model{} for 5 rounds yields only 79.0 AP, below the default \model{} (79.7 AP).
One possible explanation is that merging a larger number of decision paths into a single path may lose fine-grained splitting patterns, while the resulting longer context may distract the LLM from the tabular features.
Although this hypothesis requires systematic validation, which we leave to future work, it suggests that more boosting rounds may be needed to effectively leverage XGBoost-Large. 

\paragraph{More rounds with logit decay.}
To match the $20\times$ estimator ratio, we scale the number of boosting rounds from 5 to 100 while keeping 6 epochs per round, and train up to 30 rounds due to computational constraints.
At this scale, however, \emph{logit accumulation} becomes a practical issue: the magnitude of the accumulated logit $F_r$ (\Cref{eq:boosting}) grows unboundedly across rounds, producing overconfident predictions that resist correction in later rounds~\citep{friedman2000additive,mease2008evidence}.
Hence, we apply \emph{logit decay} with a fixed factor $\alpha < 1$:
$F_r = \alpha \cdot F_{r-1} + \eta \cdot f_{r}.$
This bounds the effective window to $\sim 1/(1-\alpha)$ rounds. We set $\alpha = 0.9$ as a conservative starting point, corresponding to an effective window of $\sim$10 rounds; finer tuning of $\alpha$ remains an open direction.

\paragraph{Results.}
\Cref{tab:scaling} summarizes the results. \model{} with XGBoost-Large, logit decay, and 30 rounds achieves 80.3 AP, surpassing both standalone XGBoost-Large (79.5 AP) and the default \model{} (79.7 AP).
Applying the same decay in the 5-round setting, however, degrades performance (78.1 vs.\ 79.7 AP), confirming that logit decay is beneficial specifically for large boosting horizons.
These results indicate \model{} generalizes to stronger tree models: pairing with XGBoost-Large and logit decay surpasses both the standalone XGBoost-Large and the default \model{}, with larger boosting horizons being key to unlocking the full benefit---a direction we leave for future work.

\section{Related Work}
\label{sec:related-work}

\paragraph{LLMs for few-shot tabular prediction.}
Recent work adapts LLMs to tabular prediction by serializing features into natural language, using either parameter-efficient fine-tuning (PEFT) or prompt-based inference.
Fine-tuning approaches include TabLLM~\citep{tabllm-2022}, which shows that careful table-to-text serialization enables competitive performance in the low-data regime, as well as UniPredict~\citep{wang2023unipredict} and TabuLa-8B~\citep{gardner2024large}, which demonstrate strong cross-dataset generalization under limited labels.
Prompt-based methods instead leverage LLM reasoning without task-specific training: LIFT~\citep{dinh2022lift} treats tabular prediction as a language task, while FeatLLM~\citep{han2024large} uses LLMs to automate feature engineering.
Prompt-to-Transfer (P2T)~\citep{nam2024tabular} enables transfer across heterogeneous tabular datasets by constructing pseudo-demonstrations for in-context learning.
DeLTa~\citep{ye2025llm} integrates LLMs with decision tree rules as intermediaries.
Despite this progress, gradient-boosted tree priors have not been explicitly incorporated into LLM training; \model{} bridges this gap.

\paragraph{Boosting for tabular prediction.}
XGBoost, LightGBM, and CatBoost~\citep{chen2016xgboost,ke2017lightgbm,prokhorenkova2018catboost} have long been state-of-the-art for tabular data.
However, their non-differentiable tree ensembles cannot directly leverage the semantic richness of unstructured text or transfer knowledge across domains.
GrowNet~\citep{badirli-2020-grownet} trains shallow networks stage-wise with residual fitting, and subsequent work extends boosting to CNNs and deep architectures~\citep{emami-2023-gbcnn-gbdnn}.
AdaBoost-style reweighting has also been adapted to deep learning~\citep{taherkhani-2020-adaboost-cnn}.
In contrast, \model{} applies boosting directly to few-shot LLM fine-tuning and integrates tree-derived decision paths as a second input view.

\paragraph{Hybrid methods combining trees and LLMs.}
One paradigm uses \emph{LLMs to enhance tree-based predictors}: LLM assists feature engineering ~\citep{han2024large,ko2025ferg,nam2024optimized,abhyankar2025llm} or produces contextual embeddings~\citep{kasneci2024enriching}, which are then fed into a downstream tree ensemble.
Another paradigm uses \emph{trees to enhance LLM-based predictors}: \citet{pattisapu2025leveraging} injects signals from tree ensembles into representation learning, DeLTa~\citep{ye2025llm} refines decision-tree rules and calibrates predictions with LLM-based corrections, and ReSS~\citep{xia2025ress} transforms decision paths into reasoning dataset for LLM fine-tuning.
\model{} falls into the second paradigm but differs in two key aspects: it integrates compressed decision paths directly into LLM fine-tuning, and it does so within a boosting framework where both views jointly contribute to residual correction at each round.
\section{Conclusion}
\label{sec:conclusion}
We presented \model{}, a framework that integrates the boosting paradigm into parameter-efficient LLM fine-tuning for few-shot tabular classification. 
By training sequential PEFT adapters as residual-correcting learners on fusion of paired feature and decision-path-informed views, \model{} reframes standard fine-tuning as structured multi-round optimization while preserving the total optimization steps. 
Our analysis shows that boosting provides consistent corrections across rounds, and that the path-informed view acts as an early-stage teacher whose influence diminishes as the LLM shifts toward feature-driven representations.
Empirically, \model{} achieves robust gains over standard fine-tuning across backbones and datasets, matches or surpasses XGBoost across shot regimes and outperforms state-of-the-art GPT-4o-based methods with a 4B model. 
We further demonstrate that the framework scales: pairing with stronger tree models and extended boosting horizons yields additional gains. 
Overall, these results suggest that boosting can serve as a general training principle for LLM fine-tuning, particularly in low-data regimes for structured data.

\section*{Limitations}
\label{sec:limitations}
\model{} has several limitations. 
First, the multi-round boosting procedure increases inference cost relative to a single-pass model (detailed in \Cref{sec:appendix-cost}).
Second, while it can optionally leverage a pre-fitted gradient boosting tree for decision paths, this introduces preprocessing overhead and partial dependence on the auxiliary model.
Third, decision-path compression is heuristic and may lose fine-grained structure as the tree ensemble grows. 
Fourth, at long boosting horizons, logit magnitudes grow under additive accumulation, requiring decay for stability; designing principled or adaptive decay schemes remains an open research direction. 
Finally, our evaluation focuses on classic tabular classification benchmarks, which may overlap with LLM pretraining corpora; generalization to regression, large-scale settings, and memorization validation remains open.

\bibliographystyle{plainnat}
\bibliography{custom.bib}

@article{mease2008evidence,
  title={Evidence Contrary to the Statistical View of Boosting},
  author={Mease, David and Wyner, Abraham},
  journal={Journal of Machine Learning Research},
  volume={9},
  pages={131--156},
  year={2008}
}

@article{friedman2000additive,
  title={Additive Logistic Regression: A Statistical View of Boosting (with Discussion and a Rejoinder by the Authors)},
  author={Friedman, Jerome and Hastie, Trevor and Tibshirani, Robert},
  journal={The Annals of Statistics},
  volume={28},
  number={2},
  pages={337--407},
  year={2000}
}

@article{ye2025llm,
  title={LLM Meeting Decision Trees on Tabular Data},
  author={Ye, Hangting and Li, Jinmeng and Zhao, He and Guo, Dandan and Chang, Yi},
  journal={arXiv preprint arXiv:2505.17918},
  year={2025}
}

@inproceedings{pattisapu2025leveraging,
  title={Leveraging structural information in tree ensembles for table representation learning},
  author={Pattisapu, Nikhil and Kasa, Siva Rajesh and Roychowdhury, Sumegh and Gupta, Karan and Bhanushali, Anish and Srinivasa Murthy, Prasanna},
  booktitle={Companion Proceedings of the ACM on Web Conference 2025},
  pages={1244--1248},
  year={2025}
}

@inproceedings{liu-2022-tfew,
  title        = {Few-Shot Parameter-Efficient Fine-Tuning is Better and Cheaper than In-Context Learning},
  author       = {Liu, Haokun and Tam, Derek and Muqeeth, Mohammed and Mohta, Jay and Huang, Tenghao and Bansal, Mohit and Raffel, Colin},
  booktitle    = {Advances in Neural Information Processing Systems (NeurIPS)},
  year         = {2022},
}

@inproceedings{tabllm-2022,
  title        = {TabLLM: Few-shot Classification of Tabular Data with Large Language Models},
  author       = {Hegselmann, Stefan and Buendia, Alejandro and Lang, Hunter and Agrawal, Monica and Jiang, Xiaoyi and Sontag, David},
  booktitle    = {International Conference on Artificial Intelligence and Statistics (AISTATS)},
  year         = {2023},
}

@article{wang2023unipredict,
  title={Unipredict: Large language models are universal tabular classifiers},
  author={Wang, Ruiyu and Wang, Zifeng and Sun, Jimeng},
  journal={arXiv preprint arXiv:2310.03266},
  year={2023}
}

@article{gardner2024large,
  title={Large scale transfer learning for tabular data via language modeling},
  author={Gardner, Josh and Perdomo, Juan C and Schmidt, Ludwig},
  journal={Advances in Neural Information Processing Systems},
  volume={37},
  pages={45155--45205},
  year={2024}
}

@article{patron2024gradient,
  title={Gradient Boosting Mapping for Dimensionality Reduction and Feature Extraction},
  author={Patron, Anri and Prasad, Ayush and Luu, Hoang Phuc Hau and Puolam{\~A}{\=I}ki, Kai},
  journal={arXiv preprint arXiv:2405.08486},
  year={2024}
}

@article{wang2015functional,
  title={Functional frank-wolfe boosting for general loss functions},
  author={Wang, Chu and Wang, Yingfei and Schapire, Robert and others},
  journal={arXiv preprint arXiv:1510.02558},
  year={2015}
}

@article{nam2024optimized,
  title={Optimized feature generation for tabular data via llms with decision tree reasoning},
  author={Nam, Jaehyun and Kim, Kyuyoung and Oh, Seunghyuk and Tack, Jihoon and Kim, Jaehyung and Shin, Jinwoo},
  journal={Advances in neural information processing systems},
  volume={37},
  pages={92352--92380},
  year={2024}
}

@inproceedings{han2024large,
  title={Large Language Models Can Automatically Engineer Features for Few-Shot Tabular Learning},
  author={Han, Sungwon and Yoon, Jinsung and Arik, Sercan O and Pfister, Tomas},
  booktitle={International Conference on Machine Learning},
  pages={17454--17479},
  year={2024},
  organization={PMLR}
}

@article{kasneci2024enriching,
  title={Enriching tabular data with contextual llm embeddings: A comprehensive ablation study for ensemble classifiers},
  author={Kasneci, Gjergji and Kasneci, Enkelejda},
  journal={arXiv preprint arXiv:2411.01645},
  year={2024}
}

@article{abhyankar2025llm,
  title={Llm-fe: Automated feature engineering for tabular data with llms as evolutionary optimizers},
  author={Abhyankar, Nikhil and Shojaee, Parshin and Reddy, Chandan K},
  journal={arXiv preprint arXiv:2503.14434},
  year={2025}
}

@inproceedings{ko2025ferg,
  title={Ferg-llm: Feature engineering by reason generation large language models},
  author={Ko, Jeonghyun and Park, Gyeongyun and Lee, Donghoon and Lee, Kyunam},
  booktitle={Findings of the Association for Computational Linguistics: NAACL 2025},
  pages={4211--4228},
  year={2025}
}

@article{one-llm-is-not-enough-2023,
  title        = {One {LLM} is not Enough: Harnessing the Power of Ensemble Learning for Medical Question Answering},
  author       = {Zhang, J. and others},
  journal      = {medRxiv},
  year         = {2023},
  url          = {https://www.medrxiv.org/content/10.1101/2023.12.21.23300380v1},
}

@article{emami-2023-gbcnn-gbdnn,
  title         = {A Gradient Boosting Approach for Training Convolutional and Deep Neural Networks},
  author        = {Emami, Seyedsaman and Mart{\'i}nez-Mu{\~n}oz, Gonzalo},
  journal       = {IEEE Open Journal of Signal Processing},
  volume        = {4},
  pages         = {313--321},
  year          = {2023},
  doi           = {10.1109/OJSP.2023.3279011},
}

@article{taherkhani-2020-adaboost-cnn,
  title         = {{A}da{B}oost-{CNN}: An adaptive boosting algorithm for convolutional neural networks to classify multi-class imbalanced datasets using transfer learning},
  author        = {Taherkhani, Aboozar and Cosma, Georgina and McGinnity, T. M.},
  journal       = {Neurocomputing},
  volume        = {404},
  pages         = {351--366},
  year          = {2020},
  doi           = {10.1016/j.neucom.2020.03.064},
}

@inproceedings{badirli-2020-grownet,
  title        = {GrowNet: Improving Deep Neural Networks with Gradient Boosting},
  author       = {Badirli, Sarkis and Liu, Tianyu and Biau, G{\'e}rard and Tan, V. Y. F.},
  booktitle    = {Advances in Neural Information Processing Systems (NeurIPS)},
  year         = {2020},
}

@inproceedings{chen2016xgboost,
  title={Xgboost: A scalable tree boosting system},
  author={Chen, Tianqi and Guestrin, Carlos},
  booktitle={Proceedings of the 22nd acm sigkdd international conference on knowledge discovery and data mining},
  pages={785--794},
  year={2016}
}

@article{zhang2025t5gemma,
  title={T5Gemma 2: Seeing, Reading, and Understanding Longer},
  author={Zhang, Biao and Suganthan, Paul and Liu, Ga{\"e}l and Philippov, Ilya and Dua, Sahil and Hora, Ben and Black, Kat and Martins, Gus and Sanseviero, Omar and Pathak, Shreya and others},
  journal={arXiv preprint arXiv:2512.14856},
  year={2025}
}

@article{yang2025qwen3,
  title={Qwen3 technical report},
  author={Yang, An and Li, Anfeng and Yang, Baosong and Zhang, Beichen and Hui, Binyuan and Zheng, Bo and Yu, Bowen and Gao, Chang and Huang, Chengen and Lv, Chenxu and others},
  journal={arXiv preprint arXiv:2505.09388},
  year={2025}
}

@article{dinh2022lift,
  title={Lift: Language-interfaced fine-tuning for non-language machine learning tasks},
  author={Dinh, Tuan and Zeng, Yuchen and Zhang, Ruisu and Lin, Ziqian and Gira, Michael and Rajput, Shashank and Sohn, Jy-yong and Papailiopoulos, Dimitris and Lee, Kangwook},
  journal={Advances in Neural Information Processing Systems},
  volume={35},
  pages={11763--11784},
  year={2022}
}

@inproceedings{nam2024tabular,
  title={Tabular Transfer Learning via Prompting LLMs},
  author={Nam, Jaehyun and Song, Woomin and Park, Seong Hyeon and Tack, Jihoon and Yun, Sukmin and Kim, Jaehyung and Oh, Kyu Hwan and Shin, Jinwoo},
  booktitle={Conference on Language Modeling (COLM)},
  year={2024}
}

@inproceedings{yan2024making,
  title={Making Pre-trained Language Models Great on Tabular Prediction},
  author={Yan, Jiahuan and Zheng, Bo and Xu, Hongxia and Zhu, Yiheng and Chen, Danny Z and Sun, Jimeng and Wu, Jian and Chen, Jintai},
  booktitle={International Conference on Learning Representations (ICLR)},
  year={2024}
}

@inproceedings{wen2024supervised,
  title={From supervised to generative: A novel paradigm for tabular deep learning with large language models},
  author={Wen, Xumeng and Zhang, Han and Zheng, Shun and Xu, Wei and Bian, Jiang},
  booktitle={Proceedings of the 30th ACM SIGKDD Conference on Knowledge Discovery and Data Mining},
  pages={3323--3333},
  year={2024}
}

@misc{yeh2008blood,
  author       = {Yeh, I-Cheng},
  title        = {Blood Transfusion Service Center},
  howpublished = {UCI Machine Learning Repository},
  year         = {2008},
  doi          = {10.24432/C5GS39},
  url          = {https://doi.org/10.24432/C5GS39}
}

@misc{hofmann1994statlog,
  author       = {Hofmann, Hans},
  title        = {Statlog (German Credit Data)},
  howpublished = {UCI Machine Learning Repository},
  year         = {1994},
  doi          = {10.24432/C5NC77},
  url          = {https://doi.org/10.24432/C5NC77}
}

@inproceedings{bohanec1988knowledge,
  author    = {Bohanec, Marko and Rajkovic, Vladislav},
  title     = {Knowledge Acquisition and Explanation for Multiattribute Decision Making},
  booktitle = {8th International Workshop on Expert Systems and their Applications},
  pages     = {59--78},
  address   = {Avignon, France},
  year      = {1988}
}

@article{moro2014bank,
  author  = {Moro, S{\'e}rgio and Cortez, Paulo and Rita, Paulo},
  title   = {A Data-Driven Approach to Predict the Success of Bank Telemarketing},
  journal = {Decision Support Systems},
  volume  = {62},
  pages   = {22--31},
  year    = {2014}
}

@inproceedings{kohavi1996scaling,
  author    = {Kohavi, Ron},
  title     = {Scaling Up the Accuracy of Naive-Bayes Classifiers: A Decision-Tree Hybrid},
  booktitle = {Proceedings of the Second International Conference on Knowledge Discovery and Data Mining (KDD)},
  pages     = {202--207},
  year      = {1996}
}

@article{liao20243,
  title={3-in-1: 2d rotary adaptation for efficient finetuning, efficient batching and composability},
  author={Liao, Baohao and Monz, Christof},
  journal={Advances in Neural Information Processing Systems},
  volume={37},
  pages={35018--35048},
  year={2024}
}

@inproceedings{
zhao2024tuning,
title={Tuning LayerNorm in Attention: Towards Efficient Multi-Modal {LLM} Finetuning},
author={Bingchen Zhao and Haoqin Tu and Chen Wei and Jieru Mei and Cihang Xie},
booktitle={The Twelfth International Conference on Learning Representations},
year={2024},
url={https://openreview.net/forum?id=YR3ETaElNK}
}

@article{gao2025optimization,
  title={Optimization-Inspired Few-Shot Adaptation for Large Language Models},
  author={Gao, Boyan and Wang, Xin and Yang, Yibo and Clifton, David},
  journal={arXiv preprint arXiv:2505.19107},
  year={2025}
}

@article{lai2025joint,
  title={Joint localization and activation editing for low-resource fine-tuning},
  author={Lai, Wen and Fraser, Alexander and Titov, Ivan},
  journal={arXiv preprint arXiv:2502.01179},
  year={2025}
}

@article{yaras2024compressible,
  title={Compressible dynamics in deep overparameterized low-rank learning \& adaptation},
  author={Yaras, Can and Wang, Peng and Balzano, Laura and Qu, Qing},
  journal={arXiv preprint arXiv:2406.04112},
  year={2024}
}

@inproceedings{hu2022lora,
  title={LoRA: Low-Rank Adaptation of Large Language Models},
  author={Hu, Edward J and Shen, Yelong and Wallis, Phillip and Allen-Zhu, Zeyuan and Li, Yuanzhi and Wang, Shean and Wang, Lu and Chen, Weizhu},
  booktitle={International Conference on Learning Representations (ICLR)},
  year={2022}
}

@article{prokhorenkova2018catboost,
  title={CatBoost: unbiased boosting with categorical features},
  author={Prokhorenkova, Liudmila and Gusev, Gleb and Vorobev, Aleksandr and Dorogush, Anna Veronika and Gulin, Andrey},
  journal={Advances in neural information processing systems},
  volume={31},
  year={2018}
}

@article{ke2017lightgbm,
  title={Lightgbm: A highly efficient gradient boosting decision tree},
  author={Ke, Guolin and Meng, Qi and Finley, Thomas and Wang, Taifeng and Chen, Wei and Ma, Weidong and Ye, Qiwei and Liu, Tie-Yan},
  journal={Advances in neural information processing systems},
  volume={30},
  year={2017}
}

@inproceedings{gorishniy2021revisiting,
  title={Revisiting Deep Learning Models for Tabular Data},
  author={Gorishniy, Yury and Rubachev, Ivan and Khrulkov, Valentin and Babenko, Artem},
  booktitle={Advances in Neural Information Processing Systems (NeurIPS)},
  year={2021}
}

@inproceedings{grinsztajn2022why,
  title={Why do tree-based models still outperform deep learning on typical tabular data?},
  author={Grinsztajn, Léo and Oyallon, Edouard and Varoquaux, Gaël},
  booktitle={Advances in Neural Information Processing Systems (NeurIPS Datasets and Benchmarks)},
  year={2022}
}

@article{ye2024closer,
  title={A closer look at deep learning methods on tabular datasets},
  author={Ye, Han-Jia and Liu, Si-Yang and Cai, Hao-Run and Zhou, Qi-Le and Zhan, De-Chuan},
  journal={arXiv preprint arXiv:2407.00956},
  year={2024}
}

@inproceedings{ma2025tabdpt,
  title={TabDPT: Scaling Tabular Foundation Models on Real Data},
  author={Ma, X. and others},
  booktitle={Advances in Neural Information Processing Systems (NeurIPS)},
  year={2025}
}

@inproceedings{tschalzev2024datacentric,
  title={A Data-Centric Perspective on Evaluating Machine Learning Models for Tabular Data},
  author={Tschalzev, Anton and others},
  booktitle={Advances in Neural Information Processing Systems (NeurIPS Datasets and Benchmarks)},
  year={2024}
}

@inproceedings{nam2023stunt,
  title={STUNT: Few-Shot Tabular Learning with Self-Generated Tasks},
  author={Nam, Joonseok and others},
  booktitle={International Conference on Learning Representations (ICLR)},
  year={2023}
}

@article{liu2024d2r2,
  title={D2r2: Diffusion-based representation with random distance matching for tabular few-shot learning},
  author={Liu, Ruoxue and Fang, Linjiajie and Wang, Wenjia and Jing, Bing-Yi},
  journal={Advances in Neural Information Processing Systems},
  volume={37},
  pages={36890--36913},
  year={2024}
}

@article{erickson2025tabarena,
  title={TabArena: A Living Benchmark for Machine Learning on Tabular Data},
  author={Erickson, Nick and Purucker, Lennart and Tschalzev, Andrej and Holzmüller, David and others},
  journal={arXiv preprint arXiv:2506.16791},
  year={2025}
}

@inproceedings{tabred2025,
  title={TabReD: Analyzing Pitfalls and Filling the Gaps in Tabular Deep Learning Benchmarks},
  author={Rubachev, Ivan and others},
  booktitle={International Conference on Learning Representations (ICLR)},
  year={2025}
}

@inproceedings{xu2007adarank,
  title={Adarank: a boosting algorithm for information retrieval},
  author={Xu, Jun and Li, Hang},
  booktitle={Proceedings of the 30th annual international ACM SIGIR conference on Research and development in information retrieval},
  pages={391--398},
  year={2007}
}

@article{van2014endgame,
  title={Endgame analysis of dou shou qi},
  author={van Rijn, Jan N and Vis, Jonathan K},
  journal={ICGA journal},
  volume={37},
  number={2},
  pages={120--124},
  year={2014},
  publisher={SAGE Publications Sage UK: London, England}
}

@article{janosi1988heart,
  title={Heart disease. UCI machine learning repository},
  author={Janosi, Andras and Steinbrunn, William and Pfisterer, Matthias and Detrano, Robert},
  journal={UCI Machine Learning Repository},
  year={1988}
}

@inproceedings{smith1988using,
  title={Using the ADAP learning algorithm to forecast the onset of diabetes mellitus},
  author={Smith, Jack W and Everhart, James E and Dickson, William C and Knowler, William C and Johannes, Robert Scott},
  booktitle={Proceedings of the annual symposium on computer application in medical care},
  pages={261},
  year={1988}
}

@inproceedings{akiba2019optuna,
  title={Optuna: A Next-generation Hyperparameter Optimization Framework},
  author={Akiba, Takuya and Sano, Shotaro and Yanase, Toshihiko and Ohta, Takeru and Koyama, Masanori},
  booktitle={Proceedings of the 25th ACM SIGKDD International Conference on Knowledge Discovery \& Data Mining},
  pages={2623--2631},
  year={2019}
}

@article{xia2025ress,
  title={ReSS: Learning Reasoning Models for Tabular Data Prediction via Symbolic Scaffold},
  author={Xia, Herun and others},
  journal={arXiv preprint arXiv:2505.00562},
  year={2025}
}

\appendix
\onecolumn

\crefname{section}{Appendix}{Appendices}
\Crefname{section}{Appendix}{Appendices}

\section{Decision-Path Compression Example}
\label[appendix]{sec:appendix-path-example}

We provide a concrete example of the decision-path compression process. Given two paths $\pi_a, \pi_b$ assigned to the same round group $\mathcal{G}_r$, the condensed path $\pi^*_r(\mathbf{x}_i)$ is obtained by intersecting constraints on overlapping features:
\begin{align*}
    \pi_a(\mathbf{x}_i) &: (18 < \texttt{age} \leq 40) \wedge (\texttt{job} = \texttt{engineer}) \\
    \pi_b(\mathbf{x}_i) &: (30 < \texttt{age} \leq 50) \wedge (\texttt{balance} > 0)
             \wedge (\texttt{job} \neq \texttt{teacher}) \\
    \pi^*_r(\mathbf{x}_i) &: (30 < \texttt{age} \leq 40) \wedge (\texttt{balance} > 0)
             \wedge (\texttt{job} = \texttt{engineer})
\end{align*}

\paragraph{Constraint merging rules.}
Since every decision path in a round group corresponds to the same sample $\mathbf{x}_i$, all paths are guaranteed to be merge-compatible: the sample's own feature values always lie within the intersection of any two constraints from different trees. The merging rules for each constraint type are:
\begin{itemize}
    \item \textbf{Numerical $\cap$ Numerical}: take the tighter range, i.e., $[\max(\ell_1, \ell_2),\, \min(u_1, u_2))$.
    \item \textbf{Categorical positive $\cap$ Categorical positive}: both must assert the same category (guaranteed for the same sample); keep the shared assertion.
    \item \textbf{Categorical positive $\cap$ Categorical negation}: the positive assertion dominates (since the sample has that value, it cannot be excluded); keep the positive assertion.
    \item \textbf{Categorical negation $\cap$ Categorical negation}: take the union of excluded categories.
\end{itemize}
The above example illustrates the numerical and categorical positive $\cap$ negation cases: the \texttt{age} ranges are intersected to $(30, 40]$, while the positive assertion $\texttt{job} = \texttt{engineer}$ absorbs the negation $\texttt{job} \neq \texttt{teacher}$.

\section{Dataset Details}
\label[appendix]{sec:appendix-datasets}

Table~\ref{tab:dataset-stats} summarizes ten tabular classification benchmarks used in our experiments. Nine datasets are binary classification tasks, and the tenth dataset: Car is a four-class multi-task classification task. All datasets exclude Jannis are sourced from the official TabLLM repository (https://github.com/clinicalml/TabLLM), while Jannis is additionally included for comparison with DeLTa~\citep{ye2025llm} and sourced from OpenML (https://www.openml.org/search?type=data\&id=44079).
For evaluation, we use 5-fold stratified cross-validation with nested train/validation/test splits.

\begin{table}[H]
  \caption{Summary statistics of the ten tabular classification benchmarks. ``Num.'' and ``Cat.'' denote the number of numerical and categorical features, respectively. ``Majority \%'' is the percentage of the most frequent class.}
  \vspace{0.5em}
  \label{tab:dataset-stats}
  \centering
  \begin{tabular}{lcccccc}
    \toprule
    \textbf{Dataset} & \textbf{Samples} & \textbf{Features} & \textbf{Num.} & \textbf{Cat.} & \textbf{\# Classes} & \textbf{Majority \%} \\
    \midrule
    Bank & 45,211 & 16 & 6 & 10 & 2 & 88.3 \\
    Blood & 748 & 4 & 4 & 0 & 2 & 76.2\\
    California & 20,640 & 8 & 8 & 0 & 2 & 50.0 \\
    Car & 1,728 & 6 & 0 & 6 & 4 & 70.0\\
    Credit-g & 1,000 & 20 & 7 & 13 & 2 & 70.0 \\
    Diabetes & 768 & 8 & 8 & 0 & 2 & 65.1 \\
    Heart & 918 & 11 & 5 & 6 & 2 & 55.3 \\
    Income & 48,842 & 12 & 4 & 8 & 2 & 76.1 \\
    Jannis & 57,580 & 54 & 54 & 0 & 2 & 50.0 \\
    Jungle & 44,819 & 6 & 6 & 0 & 2 & 51.5 \\
    \bottomrule
  \end{tabular}
\end{table}

\section{Hyperparameter Settings and Training Details}
\label[appendix]{sec:appendix-hparams}

We detail the hyperparameter configurations employed in our experiments below.

Unless stated otherwise, we follow the TabLLM~\citep{tabllm-2022} configuration as our default setting and do not perform additional per-dataset hyperparameter tuning. \Cref{tab:llm_hyperparams} outlines the primary hyperparameters used for all LLM fine-tuning tasks, which are kept consistent across all datasets.

\begin{table}[H]
    \centering
    \caption{LLM fine-tuning hyperparameters}
    \label{tab:llm_hyperparams}
    \begin{tabular}{lc}
        \toprule
        \textbf{Hyperparameter} & \textbf{Value} \\ 
        \midrule
        Learning Rate & $3 \times 10^{-3}$ \\
        Weight Decay & 0.0 \\
        Optimizer & AdamW \\
        Warmup Ratio & 0.06 \\
        Total Epochs & 30 \\
        \bottomrule
    \end{tabular}
\end{table}

For all LLM fine-tuning, we adopt RoAd~\citep{liao20243} as the parameter-efficient fine-tuning method for its strong performance with low memory usage, and additionally make all layer normalization parameters trainable, following~\citet{zhao2024tuning}.

For \model{}, the boosting learning rate is set to $\eta = 0.3$, matching the default value of XGBoost (see \Cref{sec:ablation-study}).
To ensure a fair comparison and isolated assessment of the boosting mechanism, we constrain the total budget of epochs across all rounds to be identical to standard TabLLM. Specifically, the total budget of epochs is 30 epochs, which is split evenly across 5 boosting rounds (i.e., exactly 6 epochs per round for BoostLLM, instead of 30 epochs per round). This guarantees that BoostLLM uses the identically bounded compute as the standard one-round TabLLM baseline (see \Cref{sec:ablation-study} and \Cref{sec:appendix-parameter-sensitivity} for further details).

\section{XGBoost Training Procedure}
\label[appendix]{sec:appendix-xgboost}

For the XGBoost baseline, we perform hyperparameter optimization using Optuna~\citep{akiba2019optuna} with the TPE sampler and 100 trials per test fold, and the search space is adapted from~\citet{gorishniy2021revisiting} which we provide in \Cref{tab:hyperparams}.

For XGBoost Large (Section~\ref{sec:scaling}), the search space is identical except that \texttt{n\_estimators} is expanded to UniformInt$[10, 2000]$.

We use a nested cross-validation procedure: for each of the 5 test folds, the remaining data is further split into 4 validation folds.
Optuna optimizes average precision on the sliced validation folds.
After identifying the best hyperparameters, we retrain the model on the full training set of the test fold and evaluate on the held-out test set.
This nested design prevents information leakage from validation into test evaluation.

\begin{table}[H]
    \centering
    \caption{XGBoost hyperparameter space}
    \label{tab:hyperparams}
    \begin{tabular}{lc}
        \toprule
        \textbf{Parameter} & \textbf{Distribution} \\ 
        \midrule
        \texttt{n\_estimators} & UniformInt$[10, 100]$ \\
        \texttt{max\_depth} & UniformInt$[3, 10]$ \\
        \texttt{learning\_rate} & LogUniform$[10^{-5}, 1]$ \\
        \texttt{reg\_alpha} & LogUniform$[10^{-8}, 10^{2}]$ \\
        \texttt{reg\_lambda} & LogUniform$[10^{-8}, 10^{2}]$ \\
        \texttt{gamma} & LogUniform$[10^{-8}, 10^{2}]$ \\
        \texttt{min\_child\_weight} & LogUniform$[10^{-8}, 10^{5}]$ \\
        \texttt{subsample} & Uniform$[0.5, 1]$ \\
        \texttt{colsample\_bylevel} & Uniform$[0.5, 1]$ \\
        \texttt{colsample\_bytree} & Uniform$[0.5, 1]$ \\ 
        \bottomrule
    \end{tabular}
\end{table}

\section{Per-Dataset Results for Comparison with LLM-based Few-shot Methods}
\label[appendix]{sec:appendix-per-dataset}

\Cref{tab:compare-others} reports per-dataset accuracy of \model{} alongside other LLM-based few-shot methods at 64-shot, 256-shot, and all-shot settings. All baseline numbers are from DeLTa~\citep{ye2025llm}.
Bank, Income, and Jannis are excluded from the all-shots setting due to computational cost; GTL, LIFT-ICL, and P2T are omitted from 256-shot and all-shots as these results are not reported in the corresponding papers.
\model{} achieves the highest average accuracy across all three settings and obtains the best or second-best score on the majority of individual datasets, confirming that the gains reported in the main text are not driven by a small subset of benchmarks.

\begin{table}[H]
  \caption{Test accuracy comparison with off-the-shelf LLM-based few-shot
tabular methods. All baseline numbers are from DeLTa~\citep{ye2025llm}.
GTL, LIFT-ICL, and P2T are omitted from 256-shot as these results are
not reported in the corresponding papers.
\textbf{Bold}: best; \underline{underline}: second best.}
  \vspace{0.5em}
  \label{tab:compare-others-full}
  \centering
  \resizebox{0.8\columnwidth}{!}{%
    \begin{tabular}{cc|cccccc|c}
      \toprule
      \multirow{2}{*}{\textbf{\# Shots}} & \multirow{2}{*}{\textbf{Method (LLM used)}} & \multicolumn{7}{c}{\textbf{Datasets}} \\
       & & \textbf{Bank} & \textbf{Blood} & \textbf{Car} & \textbf{Credit-g} & \textbf{Income} & \textbf{Jannis} & \textbf{Avg.} \\
      \midrule
      64 & TabLLM (T0) & 60.2 & 24.5 & 38.6 & \textbf{67.3} & \textbf{78.5} & 26.8 & 49.3 \\
      64 & LIFT (GPT-3.5) & 72.8 & 64.3 & 50.0 & 64.3 & 74.8 & 43.0 & 61.5 \\
      64 & TP-BERTa (RoBERTa) & \textbf{84.0} & 64.8 & 41.9 & 64.6 & \underline{76.1} & 41.4 & 62.1 \\
      64 & GTL (LLaMA2) & 71.9 & 61.3 & 51.2 & 63.8 & 75.0 & 41.4 & 60.8 \\
      64 & LIFT-ICL (GPT-4o) & 46.9 & 34.8 & 32.6 & 53.8 & 61.3 & 25.1 & 42.4 \\
      64 & P2T (GPT-4o) & 56.9 & 36.8 & 33.4 & 54.2 & 63.4 & 24.3 & 44.8 \\
      64 & FeatLLM (GPT-4o) & 72.3 & \underline{69.7} & 50.8 & 64.3 & 74.9 & 45.0 & 62.8 \\
      64 & DeLTa (GPT-4o) & 73.3 & \textbf{73.2} & \underline{61.5} & \underline{66.3} & 75.8 & \underline{45.7} & \underline{66.0} \\
      64 & \textbf{\model{}} (Qwen3-4B) & \underline{77.2} & 62.2 & \textbf{86.8} & \textbf{67.3} & 74.8 & \textbf{64.0} & \textbf{72.1} \\
      \midrule
      256 & TabLLM (T0) & 78.5 & 34.4 & 47.1 & \underline{69.0} & \underline{78.7} & 37.3 & 57.5 \\
      256 & LIFT (GPT-3.5) & 72.4 & 65.1 & 43.1 & 68.2 & 77.3 & 39.8 & 61.0 \\
      256 & TP-BERTa (RoBERTa) & \textbf{85.3} & 72.1 & 40.8 & 67.7 & 77.2 & 46.4 & 64.9 \\
      256 & FeatLLM (GPT-4o) & 74.0 & \underline{72.9} & 45.9 & 65.8 & 77.2 & 47.3 & 63.9 \\
      256 & DeLTa (GPT-4o) & 76.7 & \textbf{73.2} & \underline{73.6} & \textbf{71.3} & 77.8 & \underline{47.6} & \underline{70.0} \\
      256 & \textbf{\model{}} (Qwen3-4B) & \underline{81.3} & 68.6 & \textbf{95.1} & 66.7 & \textbf{78.9} & \textbf{69.1} & \textbf{76.6} \\
      \midrule
      All & TabLLM (T0) & - & 76.1 & \underline{84.5} & 73.7 & - & - & 78.1 \\
      All & LIFT (GPT-3.5) & - & 68.9 & 72.9 & 69.1 & - & - & 70.3 \\
      All & TP-BERTa (RoBERTa) & - & 76.1 & 82.6 & 73.0 & - & - & 77.2 \\
      All & FeatLLM (GPT-4o) & - & 76.8 & 58.9 & 70.1 & - & - & 68.6 \\
      All & DeLTa (GPT-4o) & - & \textbf{82.9} & 83.6 & \textbf{78.3} & - & - & \underline{81.6} \\
      All & \textbf{\model{}} (Qwen3-4B) & - & \underline{78.7} & \textbf{99.8} & \underline{75.4} & - & - & \textbf{84.6} \\
      \bottomrule
    \end{tabular}
  }
\end{table}

\section{Few-shot tabular prediction results}
\label[appendix]{sec:appendix-shot-tables}

Tables~\ref{tab:appendix-shot-tables-1} and~\ref{tab:appendix-shot-tables-2} expand upon the aggregated few-shot results presented in the main text by providing per-dataset breakdowns across all shot configurations. For almost every individual dataset and backbone combination, BoostLLM maintains a consistent lead over the standard baseline. The uniformity of these gains confirms the main conclusion that the proposed boosting strategy generalizes effectively and provides a reliable enhancement without relying on dataset-specific conditions.

\newcommand{\shotheader}{%
  \textbf{Model} & \textbf{Method} & \textbf{Avg.} & \textbf{bank} & \textbf{blood} & \textbf{calhousing} & \textbf{Car} & \textbf{credit-g} & \textbf{diabetes} & \textbf{heart} & \textbf{income} & \textbf{jungle} \\
}

\begin{table}[H]
  \caption{Few-shot tabular prediction results (4, 8, 16, 32, 64 shots).}
  \label{tab:appendix-shot-tables-1}
  \centering
  \resizebox{\textwidth}{!}{%
  \begin{tabular}{lllccccccccc}
    \toprule
    \multicolumn{12}{c}{\textbf{4 shots}} \\
    \midrule
    \shotheader
    \midrule
    XGBoost & & 57.9 & 54.1 ± 1.4 & 53.4 ± 6.9 & 63.1 ± 9.3 & 35.4 ± 5.2 & 55.5 ± 5.4 & 54.5 ± 6.7 & 78.7 ± 7.4 & 64.2 ± 3.7 & 62.1 ± 2.7 \\
    Qwen3-4B & TabLLM   & 60.0 & 54.6 $\pm$ 3.7 & 53.2 $\pm$ 6.0 & 61.6 $\pm$ 5.2 & 43.8 $\pm$ 6.3 & 58.6 $\pm$ 7.2 & 61.4 $\pm$ 7.4 & 78.1 $\pm$ 5.4 & 66.9 $\pm$ 7.1 & 61.4 $\pm$ 5.1 \\
    & BoostLLM & 59.9 & 58.0 $\pm$ 3.8 & 53.6 $\pm$ 6.6 & 66.2 $\pm$ 11.0 & 45.7 $\pm$ 8.7 & 55.3 $\pm$ 10.2 & 72.3 $\pm$ 4.4 & 67.6 $\pm$ 11.6 & 69.4 $\pm$ 5.4 & 51.4 $\pm$ 6.6 \\
    Qwen3-8B & TabLLM   & 58.5 & 54.1 $\pm$ 1.9 & 54.6 $\pm$ 5.9 & 58.0 $\pm$ 8.3 & 39.9 $\pm$ 2.9 & 57.7 $\pm$ 7.8 & 60.9 $\pm$ 10.9 & 76.6 $\pm$ 4.9 & 65.9 $\pm$ 5.7 & 58.3 $\pm$ 4.4 \\
    & BoostLLM & 62.4 & 56.4 $\pm$ 3.4 & 55.7 $\pm$ 8.5 & 61.5 $\pm$ 8.4 & 49.8 $\pm$ 9.4 & 58.4 $\pm$ 7.7 & 60.6 $\pm$ 9.2 & 82.1 $\pm$ 7.9 & 69.2 $\pm$ 9.2 & 67.4 $\pm$ 4.4 \\
    T5Gemma2-2b & TabLLM   & 57.6 & 54.4 $\pm$ 2.2 & 50.5 $\pm$ 5.1 & 54.6 $\pm$ 5.4 & 49.1 $\pm$ 7.4 & 54.4 $\pm$ 3.0 & 57.4 $\pm$ 5.6 & 77.9 $\pm$ 8.4 & 66.2 $\pm$ 5.1 & 54.2 $\pm$ 7.0 \\
    & BoostLLM & 59.6 & 57.2 $\pm$ 1.1 & 53.2 $\pm$ 5.5 & 63.9 $\pm$ 7.6 & 53.0 $\pm$ 6.0 & 57.3 $\pm$ 6.6 & 59.8 $\pm$ 7.9 & 76.9 $\pm$ 11.1 & 61.0 $\pm$ 5.2 & 54.0 $\pm$ 4.3 \\
    T5Gemma2-8b & TabLLM   & 59.4 & 54.1 $\pm$ 2.8 & 51.2 $\pm$ 2.5 & 62.6 $\pm$ 7.0 & 52.7 $\pm$ 8.4 & 53.3 $\pm$ 3.3 & 57.1 $\pm$ 6.0 & 78.6 $\pm$ 5.9 & 67.1 $\pm$ 3.2 & 58.1 $\pm$ 5.9 \\
    & BoostLLM & 60.0 & 55.2 $\pm$ 1.9 & 53.1 $\pm$ 6.3 & 55.5 $\pm$ 5.4 & 54.1 $\pm$ 3.4 & 58.9 $\pm$ 4.5 & 64.5 $\pm$ 11.2 & 78.3 $\pm$ 5.2 & 65.4 $\pm$ 4.4 & 55.2 $\pm$ 5.5 \\
    \midrule
    \multicolumn{12}{c}{\textbf{8 shots}} \\
    \midrule
    \shotheader
    \midrule
    XGBoost & & 59.9 & 58.6 ± 3.8 & 56.9 ± 7.2 & 65.0 ± 7.7 & 39.5 ± 4.9 & 58.2 ± 2.2 & 53.0 ± 13.4 & 80.3 ± 14.1 & 68.4 ± 3.4 & 59.3 ± 6.9\\
    Qwen3-4B & TabLLM   & 62.1 & 54.6 $\pm$ 2.5 & 57.7 $\pm$ 2.8 & 64.1 $\pm$ 6.5 & 47.9 $\pm$ 7.1 & 61.1 $\pm$ 5.2 & 62.8 $\pm$ 5.6 & 80.4 $\pm$ 11.6 & 67.4 $\pm$ 3.5 & 63.3 $\pm$ 3.7 \\
    & BoostLLM & 63.0 & 58.2 $\pm$ 6.4 & 55.3 $\pm$ 5.5 & 65.6 $\pm$ 5.2 & 46.2 $\pm$ 3.6 & 61.9 $\pm$ 4.7 & 63.7 $\pm$ 10.4 & 80.7 $\pm$ 10.7 & 77.5 $\pm$ 3.1 & 57.6 $\pm$ 6.9 \\
    Qwen3-8B & TabLLM   & 61.1 & 56.1 $\pm$ 5.3 & 55.1 $\pm$ 5.2 & 64.2 $\pm$ 6.1 & 44.2 $\pm$ 13.1 & 58.8 $\pm$ 9.5 & 63.7 $\pm$ 4.7 & 81.2 $\pm$ 6.4 & 68.6 $\pm$ 5.2 & 58.0 $\pm$ 5.3 \\
    & BoostLLM & 64.1 & 62.6 $\pm$ 6.8 & 60.7 $\pm$ 8.3 & 62.1 $\pm$ 2.5 & 50.4 $\pm$ 3.8 & 60.4 $\pm$ 4.9 & 57.8 $\pm$ 12.3 & 86.3 $\pm$ 5.9 & 75.7 $\pm$ 4.5 & 61.3 $\pm$ 5.7 \\
    T5Gemma2-2b & TabLLM   & 61.2 & 55.8 $\pm$ 1.5 & 53.5 $\pm$ 7.3 & 61.3 $\pm$ 9.7 & 55.7 $\pm$ 9.4 & 61.3 $\pm$ 4.8 & 54.9 $\pm$ 8.3 & 78.5 $\pm$ 8.4 & 71.4 $\pm$ 5.3 & 58.3 $\pm$ 5.2 \\
    & BoostLLM & 60.8 & 58.8 $\pm$ 4.0 & 54.8 $\pm$ 3.5 & 64.5 $\pm$ 4.9 & 54.6 $\pm$ 9.6 & 60.0 $\pm$ 4.6 & 55.5 $\pm$ 13.7 & 80.8 $\pm$ 11.4 & 63.1 $\pm$ 8.4 & 55.1 $\pm$ 8.7 \\
    T5Gemma2-8b & TabLLM   & 60.4 & 56.0 $\pm$ 1.4 & 57.6 $\pm$ 4.5 & 60.7 $\pm$ 3.8 & 52.2 $\pm$ 9.2 & 60.8 $\pm$ 3.7 & 58.0 $\pm$ 4.0 & 76.3 $\pm$ 6.9 & 65.4 $\pm$ 4.1 & 56.5 $\pm$ 5.8 \\
    & BoostLLM & 61.9 & 56.9 $\pm$ 2.0 & 56.8 $\pm$ 8.1 & 60.2 $\pm$ 3.2 & 56.8 $\pm$ 5.2 & 61.5 $\pm$ 3.5 & 58.4 $\pm$ 15.9 & 82.2 $\pm$ 7.8 & 71.2 $\pm$ 5.9 & 53.4 $\pm$ 3.9 \\
    \midrule
    \multicolumn{12}{c}{\textbf{16 shots}} \\
    \midrule
    \shotheader
    \midrule
    XGBoost & & 65.5 & 60.9 $\pm$ 1.7 & 57.4 $\pm$ 5.2 & 70.9 $\pm$ 5.6 & 42.9 $\pm$ 1.8 & 60.2 $\pm$ 4.6 & 70.4 $\pm$ 3.6 & 87.0 $\pm$ 2.2 & 71.8 $\pm$ 2.2 & 68.4 $\pm$ 4.3 \\
    Qwen3-4B & TabLLM & 63.2 & 54.1 $\pm$ 1.9 & 56.9 $\pm$ 5.5 & 63.3 $\pm$ 3.6 & 55.9 $\pm$ 11.7 & 59.0 $\pm$ 7.5 & 63.0 $\pm$ 6.6 & 81.2 $\pm$ 6.0 & 70.0 $\pm$ 2.9 & 64.9 $\pm$ 4.8 \\
     & BoostLLM & 68.2 & 60.7 $\pm$ 2.1 & 57.3 $\pm$ 7.2 & 75.3 $\pm$ 6.6 & 57.0 $\pm$ 8.1 & 61.0 $\pm$ 8.0 & 72.3 $\pm$ 5.1 & 85.6 $\pm$ 2.0 & 78.5 $\pm$ 3.0 & 66.4 $\pm$ 3.5 \\
    Qwen3-8B & TabLLM & 63.3 & 57.3 $\pm$ 3.6 & 57.4 $\pm$ 4.4 & 61.2 $\pm$ 7.3 & 57.8 $\pm$ 7.5 & 56.6 $\pm$ 6.4 & 64.9 $\pm$ 7.8 & 79.1 $\pm$ 4.9 & 70.3 $\pm$ 4.6 & 64.8 $\pm$ 6.6 \\
     & BoostLLM & 68.5 & 63.7 $\pm$ 1.2 & 56.6 $\pm$ 4.6 & 70.5 $\pm$ 3.1 & 59.3 $\pm$ 10.1 & 62.5 $\pm$ 7.5 & 71.5 $\pm$ 7.5 & 87.5 $\pm$ 1.3 & 75.0 $\pm$ 3.1 & 69.9 $\pm$ 1.8 \\
    T5Gemma2-2B & TabLLM & 64.3 & 55.5 $\pm$ 2.4 & 56.8 $\pm$ 4.5 & 62.9 $\pm$ 7.7 & 63.4 $\pm$ 5.7 & 59.0 $\pm$ 6.6 & 62.8 $\pm$ 8.5 & 78.9 $\pm$ 5.7 & 74.5 $\pm$ 2.6 & 64.7 $\pm$ 4.8 \\
     & BoostLLM & 66.3 & 60.7 $\pm$ 3.8 & 57.8 $\pm$ 5.2 & 67.3 $\pm$ 8.2 & 59.4 $\pm$ 6.1 & 59.3 $\pm$ 7.1 & 70.0 $\pm$ 6.2 & 87.3 $\pm$ 3.4 & 73.9 $\pm$ 1.0 & 61.5 $\pm$ 5.1 \\
    T5Gemma2-8B & TabLLM & 64.3 & 54.8 $\pm$ 2.4 & 58.1 $\pm$ 3.6 & 66.4 $\pm$ 7.5 & 62.3 $\pm$ 6.4 & 57.9 $\pm$ 7.6 & 64.0 $\pm$ 2.5 & 81.0 $\pm$ 5.1 & 72.2 $\pm$ 2.7 & 61.9 $\pm$ 3.6 \\
     & BoostLLM & 66.5 & 60.8 $\pm$ 3.8 & 54.0 $\pm$ 7.2 & 67.3 $\pm$ 5.7 & 59.2 $\pm$ 8.0 & 60.0 $\pm$ 6.2 & 72.5 $\pm$ 4.4 & 87.4 $\pm$ 0.9 & 77.9 $\pm$ 2.6 & 59.4 $\pm$ 6.2 \\
    \midrule
    \multicolumn{12}{c}{\textbf{32 shots}} \\
    \midrule
    \shotheader
    \midrule
    XGBoost &  & 70.4 & 65.3 $\pm$ 1.0 & 61.1 $\pm$ 6.5 & 74.2 $\pm$ 5.2 & 59.3 $\pm$ 7.0 & 63.3 $\pm$ 5.3 & 76.9 $\pm$ 5.2 & 82.8 $\pm$ 3.5 & 74.1 $\pm$ 2.6 & 77.0 $\pm$ 2.8 \\
    Qwen3-4B & TabLLM & 67.0 & 54.8 $\pm$ 2.2 & 60.3 $\pm$ 6.3 & 68.6 $\pm$ 1.7 & 72.1 $\pm$ 11.1 & 59.1 $\pm$ 3.1 & 67.3 $\pm$ 9.3 & 78.4 $\pm$ 3.0 & 72.3 $\pm$ 1.3 & 69.8 $\pm$ 2.9 \\
     & BoostLLM & 72.8 & 62.8 $\pm$ 4.9 & 63.5 $\pm$ 5.0 & 75.6 $\pm$ 4.0 & 72.3 $\pm$ 5.7 & 62.6 $\pm$ 4.6 & 77.8 $\pm$ 4.3 & 86.2 $\pm$ 1.6 & 77.5 $\pm$ 0.7 & 76.6 $\pm$ 1.9 \\
    Qwen3-8B & TabLLM & 67.3 & 57.6 $\pm$ 4.5 & 59.5 $\pm$ 5.2 & 61.8 $\pm$ 6.7 & 74.5 $\pm$ 9.7 & 61.4 $\pm$ 4.5 & 70.5 $\pm$ 5.5 & 78.6 $\pm$ 5.1 & 71.9 $\pm$ 1.9 & 69.5 $\pm$ 3.7 \\
     & BoostLLM & 72.8  & 65.6 $\pm$ 1.2 & 64.4 $\pm$ 6.3 & 72.2 $\pm$ 5.9 & 77.1 $\pm$ 5.4 & 61.2 $\pm$ 3.8 & 78.2 $\pm$ 4.3 & 84.0 $\pm$ 4.7 & 76.2 $\pm$ 2.3 & 76.4 $\pm$ 2.4 \\
    T5Gemma2-2B & TabLLM & 68.8 & 58.5 $\pm$ 2.5 & 60.6 $\pm$ 8.4 & 63.0 $\pm$ 8.5 & 75.1 $\pm$ 7.0 & 60.6 $\pm$ 3.7 & 73.9 $\pm$ 5.1 & 81.1 $\pm$ 6.4 & 73.8 $\pm$ 0.8 & 72.5 $\pm$ 3.9 \\
     & BoostLLM & 70.1 & 64.9 $\pm$ 2.1 & 59.9 $\pm$ 5.8 & 67.9 $\pm$ 6.3 & 69.4 $\pm$ 9.7 & 62.1 $\pm$ 4.5 & 78.6 $\pm$ 5.7 & 83.8 $\pm$ 8.1 & 76.1 $\pm$ 2.0 & 68.0 $\pm$ 5.6 \\
    T5Gemma2-8B & TabLLM & 67.8 & 56.3 $\pm$ 2.8 & 59.7 $\pm$ 5.3 & 66.3 $\pm$ 11.2 & 73.3 $\pm$ 6.2 & 58.7 $\pm$ 2.7 & 70.5 $\pm$ 4.3 & 81.5 $\pm$ 7.9 & 73.5 $\pm$ 3.7 & 70.4 $\pm$ 3.7 \\
     & BoostLLM & 71.0 & 62.7 $\pm$ 1.4 & 61.5 $\pm$ 6.6 & 70.2 $\pm$ 6.2 & 70.5 $\pm$ 5.6 & 61.8 $\pm$ 4.8 & 77.8 $\pm$ 6.5 & 84.9 $\pm$ 6.2 & 78.8 $\pm$ 1.6 & 70.4 $\pm$ 6.4 \\
    \midrule
    \multicolumn{12}{c}{\textbf{64 shots}} \\
    \midrule
    \shotheader
    \midrule
    XGBoost &  & 75.4 & 66.8 $\pm$ 0.9 & 68.6 $\pm$ 5.1 & 80.2 $\pm$ 1.6 & 72.5 $\pm$ 9.3 & 66.8 $\pm$ 5.8 & 74.1 $\pm$ 1.3 & 89.2 $\pm$ 4.1 & 78.5 $\pm$ 1.8 & 82.0 $\pm$ 1.0 \\
    Qwen3-4B & TabLLM & 71.7 & 61.5 $\pm$ 3.9 & 60.6 $\pm$ 4.2 & 76.9 $\pm$ 4.4 & 80.1 $\pm$ 6.5 & 60.6 $\pm$ 5.4 & 70.5 $\pm$ 5.3 & 87.1 $\pm$ 4.9 & 75.5 $\pm$ 2.5 & 72.8 $\pm$ 2.4 \\
     & BoostLLM & 77.0 & 67.9 $\pm$ 1.5 & 68.2 $\pm$ 5.1 & 80.7 $\pm$ 2.3 & 85.4 $\pm$ 2.6 & 66.2 $\pm$ 3.5 & 75.4 $\pm$ 2.8 & 89.7 $\pm$ 2.5 & 79.8 $\pm$ 2.9 & 80.2 $\pm$ 1.8 \\
    Qwen3-8B & TabLLM & 70.1 & 62.3 $\pm$ 5.9 & 62.8 $\pm$ 2.9 & 70.1 $\pm$ 9.0 & 78.1 $\pm$ 9.6 & 60.3 $\pm$ 3.7 & 69.8 $\pm$ 3.8 & 85.8 $\pm$ 4.7 & 73.8 $\pm$ 4.8 & 67.9 $\pm$ 10.5 \\
     & BoostLLM & 76.9 & 69.5 $\pm$ 0.7 & 68.6 $\pm$ 6.3 & 77.9 $\pm$ 1.8 & 84.9 $\pm$ 4.3 & 65.6 $\pm$ 4.1 & 75.8 $\pm$ 2.4 & 89.5 $\pm$ 3.1 & 80.1 $\pm$ 2.5 & 80.1 $\pm$ 2.2 \\
    T5Gemma2-2B & TabLLM & 71.0 & 61.1 $\pm$ 2.7 & 62.3 $\pm$ 5.3 & 73.4 $\pm$ 2.1 & 82.3 $\pm$ 4.1 & 62.4 $\pm$ 2.8 & 62.6 $\pm$ 4.8 & 86.0 $\pm$ 4.1 & 74.4 $\pm$ 2.7 & 74.7 $\pm$ 4.3 \\
     & BoostLLM & 75.4 & 67.6 $\pm$ 1.3 & 67.5 $\pm$ 4.7 & 76.5 $\pm$ 2.0 & 81.7 $\pm$ 6.5 & 64.2 $\pm$ 4.0 & 76.3 $\pm$ 2.9 & 90.7 $\pm$ 2.9 & 78.9 $\pm$ 2.8 & 75.3 $\pm$ 1.1 \\
    T5Gemma2-8B & TabLLM & 71.9 & 62.6 $\pm$ 2.2 & 61.9 $\pm$ 5.5 & 74.4 $\pm$ 1.4 & 80.9 $\pm$ 7.2 & 61.9 $\pm$ 4.0 & 71.7 $\pm$ 6.0 & 85.9 $\pm$ 4.8 & 74.2 $\pm$ 4.2 & 73.6 $\pm$ 2.4 \\
     & BoostLLM & 76.4 & 67.1 $\pm$ 1.4 & 67.9 $\pm$ 4.6 & 76.6 $\pm$ 3.7 & 86.1 $\pm$ 1.7 & 63.8 $\pm$ 4.7 & 76.3 $\pm$ 3.3 & 90.3 $\pm$ 2.4 & 80.5 $\pm$ 1.7 & 79.0 $\pm$ 0.9 \\
     \bottomrule
  \end{tabular}
  }
\end{table}

\begin{table}[H]
  \caption{Few-shot tabular prediction results (128, 256, 512, ALL shots).}
  \label{tab:appendix-shot-tables-2}
  \centering
  \resizebox{\textwidth}{!}{%
  \begin{tabular}{lllccccccccc}
    \toprule
    \multicolumn{12}{c}{\textbf{128 shots}} \\
    \midrule
    \shotheader
    \midrule
    XGBoost &  & 78.5 & 70.3 $\pm$ 3.4 & 66.1 $\pm$ 2.4 & 82.8 $\pm$ 1.6 & 85.9 $\pm$ 5.8 & 71.1 $\pm$ 0.7 & 76.6 $\pm$ 4.1 & 91.0 $\pm$ 2.6 & 78.9 $\pm$ 1.5 & 83.6 $\pm$ 1.8 \\
    Qwen3-4B & TabLLM & 74.8 & 65.9 $\pm$ 4.0 & 58.6 $\pm$ 6.1 & 75.7 $\pm$ 7.1 & 90.5 $\pm$ 10.4 & 68.4 $\pm$ 2.9 & 72.7 $\pm$ 3.2 & 87.5 $\pm$ 3.6 & 75.4 $\pm$ 2.5 & 78.9 $\pm$ 3.7 \\
     & BoostLLM & 79.7 & 71.3 $\pm$ 2.2 & 65.8 $\pm$ 2.8 & 84.7 $\pm$ 1.6 & 91.4 $\pm$ 5.6 & 71.3 $\pm$ 1.0 & 78.3 $\pm$ 2.8 & 91.3 $\pm$ 2.4 & 82.1 $\pm$ 1.4 & 81.2 $\pm$ 1.6 \\
    Qwen3-8B & TabLLM & 73.0 & 65.4 $\pm$ 2.0 & 62.4 $\pm$ 6.8 & 70.0 $\pm$ 5.0 & 89.0 $\pm$ 4.8 & 61.4 $\pm$ 1.5 & 71.0 $\pm$ 3.6 & 85.8 $\pm$ 3.7 & 77.2 $\pm$ 3.1 & 74.3 $\pm$ 5.0 \\
     & BoostLLM & 79.2 & 72.5 $\pm$ 1.9 & 65.3 $\pm$ 4.4 & 81.0 $\pm$ 3.1 & 90.4 $\pm$ 6.3 & 70.4 $\pm$ 0.6 & 78.9 $\pm$ 2.6 & 91.7 $\pm$ 1.8 & 81.8 $\pm$ 1.0 & 81.0 $\pm$ 1.3 \\
    T5Gemma2-2B & TabLLM & 74.2 & 66.8 $\pm$ 2.9 & 58.9 $\pm$ 6.1 & 72.8 $\pm$ 5.4 & 89.5 $\pm$ 8.2 & 67.9 $\pm$ 3.7 & 72.2 $\pm$ 2.4 & 86.7 $\pm$ 2.9 & 75.7 $\pm$ 3.5 & 77.4 $\pm$ 2.7 \\
     & BoostLLM & 79.1 & 70.5 $\pm$ 2.9 & 66.7 $\pm$ 4.3 & 81.9 $\pm$ 1.2 & 89.8 $\pm$ 7.8 & 70.6 $\pm$ 1.6 & 78.9 $\pm$ 3.7 & 91.5 $\pm$ 2.0 & 81.2 $\pm$ 1.6 & 80.5 $\pm$ 1.4 \\
    T5Gemma2-8B & TabLLM & 74.6 & 67.0 $\pm$ 2.4 & 59.3 $\pm$ 3.0 & 77.5 $\pm$ 4.2 & 90.0 $\pm$ 10.8 & 65.0 $\pm$ 4.4 & 73.1 $\pm$ 4.0 & 86.6 $\pm$ 3.7 & 75.6 $\pm$ 2.8 & 77.4 $\pm$ 1.8 \\
     & BoostLLM & 79.1 & 71.4 $\pm$ 2.0 & 65.3 $\pm$ 4.6 & 82.4 $\pm$ 2.6 & 90.9 $\pm$ 6.5 & 70.1 $\pm$ 1.3 & 77.8 $\pm$ 3.0 & 91.2 $\pm$ 1.6 & 81.3 $\pm$ 1.3 & 81.3 $\pm$ 1.5 \\
    \midrule
    \multicolumn{12}{c}{\textbf{256 shots}} \\
    \midrule
    \shotheader
    \midrule
    XGBoost &  & 81.3 & 73.3 $\pm$ 2.3 & 67.9 $\pm$ 5.8 & 85.8 $\pm$ 1.4 & 91.2 $\pm$ 3.2 & 74.3 $\pm$ 2.3 & 78.0 $\pm$ 5.4 & 91.1 $\pm$ 2.2 & 83.3 $\pm$ 0.8 & 86.9 $\pm$ 1.0 \\
    Qwen3-4B & TabLLM & 75.7 & 65.2 $\pm$ 1.7 & 62.4 $\pm$ 6.4 & 81.2 $\pm$ 4.1 & 97.9 $\pm$ 1.0 & 66.4 $\pm$ 4.5 & 68.7 $\pm$ 11.4 & 85.7 $\pm$ 1.3 & 77.3 $\pm$ 2.0 & 76.5 $\pm$ 14.3 \\
     & BoostLLM & 82.2 & 74.2 $\pm$ 1.8 & 67.8 $\pm$ 4.3 & 87.4 $\pm$ 0.8 & 96.4 $\pm$ 1.3 & 73.2 $\pm$ 1.6 & 79.5 $\pm$ 3.9 & 92.3 $\pm$ 1.2 & 84.0 $\pm$ 0.4 & 84.9 $\pm$ 0.4 \\
    Qwen3-8B & TabLLM & 75.2 & 64.9 $\pm$ 1.9 & 62.7 $\pm$ 8.9 & 77.7 $\pm$ 2.6 & 94.1 $\pm$ 6.8 & 63.6 $\pm$ 3.0 & 70.6 $\pm$ 8.6 & 86.3 $\pm$ 2.0 & 75.8 $\pm$ 2.6 & 81.5 $\pm$ 2.7 \\
     & BoostLLM & 81.5 & 74.1 $\pm$ 1.0 & 67.7 $\pm$ 4.1 & 84.7 $\pm$ 0.7 & 96.5 $\pm$ 1.5 & 72.2 $\pm$ 1.8 & 79.1 $\pm$ 3.5 & 91.9 $\pm$ 1.5 & 82.5 $\pm$ 1.5 & 84.7 $\pm$ 1.5 \\
    T5Gemma2-2B & TabLLM & 76.6 & 69.7 $\pm$ 2.4 & 62.2 $\pm$ 4.8 & 72.3 $\pm$ 4.9 & 99.0 $\pm$ 0.8 & 68.5 $\pm$ 2.3 & 73.3 $\pm$ 4.0 & 87.0 $\pm$ 2.4 & 76.0 $\pm$ 0.7 & 81.0 $\pm$ 1.7 \\
     & BoostLLM & 81.7 & 73.6 $\pm$ 1.9 & 67.2 $\pm$ 3.7 & 84.6 $\pm$ 1.0 & 97.3 $\pm$ 0.9 & 72.1 $\pm$ 1.4 & 79.0 $\pm$ 3.4 & 92.3 $\pm$ 1.6 & 83.5 $\pm$ 0.8 & 85.9 $\pm$ 1.3 \\
    T5Gemma2-8B & TabLLM & 77.2 & 69.5 $\pm$ 1.5 & 61.6 $\pm$ 3.7 & 79.3 $\pm$ 3.6 & 99.1 $\pm$ 0.6 & 65.8 $\pm$ 3.0 & 73.2 $\pm$ 5.6 & 86.9 $\pm$ 2.2 & 77.2 $\pm$ 1.8 & 82.0 $\pm$ 1.5 \\
     & BoostLLM & 81.6 & 73.9 $\pm$ 2.1 & 66.9 $\pm$ 4.4 & 85.4 $\pm$ 1.0 & 97.4 $\pm$ 1.1 & 71.0 $\pm$ 1.7 & 79.4 $\pm$ 3.4 & 92.0 $\pm$ 1.2 & 83.3 $\pm$ 1.2 & 84.7 $\pm$ 1.0 \\
    \midrule
    \multicolumn{12}{c}{\textbf{512 shots}} \\
    \midrule
    \shotheader
    \midrule
    XGBoost &  & 83.0 & 74.2 $\pm$ 1.4 & 68.4 $\pm$ 4.1 & 87.8 $\pm$ 0.5 & 96.2 $\pm$ 2.0 & 74.7 $\pm$ 3.4 & 79.4 $\pm$ 2.3 & 92.0 $\pm$ 1.8 & 84.6 $\pm$ 1.5 & 89.8 $\pm$ 0.4 \\
    Qwen3-4B & TabLLM & 78.4 & 68.4 $\pm$ 1.7 & 64.6 $\pm$ 4.4 & 83.2 $\pm$ 1.5 & 99.2 $\pm$ 1.0 & 69.4 $\pm$ 1.5 & 69.6 $\pm$ 1.3 & 88.0 $\pm$ 2.8 & 76.3 $\pm$ 3.8 & 86.9 $\pm$ 1.9 \\
     & BoostLLM & 83.4 & 75.1 $\pm$ 1.1 & 69.5 $\pm$ 3.6 & 88.5 $\pm$ 0.4 & 98.7 $\pm$ 0.8 & 73.2 $\pm$ 2.5 & 80.3 $\pm$ 3.2 & 93.4 $\pm$ 1.6 & 84.2 $\pm$ 1.8 & 88.1 $\pm$ 1.6 \\
    Qwen3-8B & TabLLM & 77.7 & 68.2 $\pm$ 2.7 & 63.7 $\pm$ 2.0 & 78.1 $\pm$ 4.2 & 99.4 $\pm$ 0.5 & 63.0 $\pm$ 2.9 & 73.1 $\pm$ 5.0 & 90.0 $\pm$ 1.3 & 77.6 $\pm$ 1.1 & 86.4 $\pm$ 0.9 \\
     & BoostLLM & 82.3 & 74.7 $\pm$ 1.2 & 68.8 $\pm$ 3.0 & 87.0 $\pm$ 1.9 & 96.6 $\pm$ 1.9 & 72.0 $\pm$ 2.4 & 79.6 $\pm$ 2.2 & 92.8 $\pm$ 1.3 & 83.5 $\pm$ 1.6 & 85.2 $\pm$ 2.2\\
    T5Gemma2-2B & TabLLM & 77.3 & 68.7 $\pm$ 1.3 & 60.6 $\pm$ 3.3 & 74.4 $\pm$ 2.0 & 99.8 $\pm$ 0.4 & 66.8 $\pm$ 2.9 & 71.5 $\pm$ 6.8 & 89.9 $\pm$ 2.2 & 78.0 $\pm$ 1.0 & 85.9 $\pm$ 0.9 \\
     & BoostLLM & 83.5 & 76.1 $\pm$ 1.6 & 70.1 $\pm$ 3.5 & 86.7 $\pm$ 1.3 & 99.1 $\pm$ 0.6 & 72.9 $\pm$ 2.8 & 79.8 $\pm$ 2.6 & 92.9 $\pm$ 1.1 & 84.6 $\pm$ 1.4 & 89.6 $\pm$ 1.2 \\
    T5Gemma2-8B & TabLLM & 77.6 & 69.1 $\pm$ 1.9 & 60.0 $\pm$ 4.0 & 78.8 $\pm$ 1.1 & 99.7 $\pm$ 0.3 & 65.7 $\pm$ 4.5 & 71.6 $\pm$ 4.7 & 88.0 $\pm$ 2.3 & 79.0 $\pm$ 1.0 & 86.9 $\pm$ 1.5 \\
     & BoostLLM & 83.3 & 75.3 $\pm$ 1.1 & 70.3 $\pm$ 2.7 & 87.4 $\pm$ 0.9 & 99.0 $\pm$ 1.1 & 72.0 $\pm$ 3.3 & 80.3 $\pm$ 2.5 & 92.8 $\pm$ 1.7 & 84.8 $\pm$ 1.5 & 88.2 $\pm$ 1.4 \\
    \midrule
    \multicolumn{12}{c}{\textbf{All shots}} \\
    \midrule
    \shotheader
    \midrule
    XGBoost &  & 83.0 & - & 68.9 $\pm$ 8.6 & - & 99.2 $\pm$ 0.6 & 76.0 $\pm$ 3.5 & 80.1 $\pm$ 4.4 & 90.8 $\pm$ 1.8 & - & - \\
    Qwen3-4B & TabLLM & 77.6 & - & 59.1 $\pm$ 3.9 & - & 100.0 $\pm$ 0.0 & 66.9 $\pm$ 3.4 & 72.0 $\pm$ 4.1 & 89.7 $\pm$ 3.4 & - & - \\
     & BoostLLM & 83.0 & - & 67.8 $\pm$ 7.2 & - & 99.9 $\pm$ 0.1 & 74.0 $\pm$ 4.1 & 80.1 $\pm$ 4.0 & 93.3 $\pm$ 1.0 & - & - \\
    Qwen3-8B & TabLLM & 78.0 & - & 64.0 $\pm$ 7.8 & - & 100.0 $\pm$ 0.0 & 65.0 $\pm$ 5.1 & 71.3 $\pm$ 5.0 & 89.5 $\pm$ 2.8 & - & - \\
     & BoostLLM & 82.6 & - & 67.4 $\pm$ 6.7 & - & 99.8 $\pm$ 0.3 & 74.0 $\pm$ 4.7 & 78.7 $\pm$ 5.5 & 93.2 $\pm$ 1.0 & - & - \\
    T5Gemma2-2B & TabLLM & 76.9 & - & 55.9 $\pm$ 3.7 & - & 100.0 $\pm$ 0.0 & 67.4 $\pm$ 3.0 & 73.3 $\pm$ 3.6 & 87.9 $\pm$ 2.4 & - & - \\
     & BoostLLM & 83.3 & - & 69.1 $\pm$ 7.3 & - & 100.0 $\pm$ 0.0 & 73.7 $\pm$ 4.6 & 80.7 $\pm$ 3.8 & 93.1 $\pm$ 1.4 & - & - \\
    T5Gemma2-8B & TabLLM & 77.4 & - & 58.3 $\pm$ 4.7 & - & 100.0 $\pm$ 0.0 & 67.8 $\pm$ 4.5 & 72.0 $\pm$ 4.0 & 88.9 $\pm$ 2.4 & - & - \\
     & BoostLLM & 83.0 & - & 68.2 $\pm$ 6.6 & - & 100.0 $\pm$ 0.0 & 72.9 $\pm$ 4.9 & 80.9 $\pm$ 4.1 & 93.0 $\pm$ 1.2 & - & - \\
    \bottomrule
  \end{tabular}
  }
\end{table}

\section{Parameter Sensitivity}
\label[appendix]{sec:appendix-parameter-sensitivity}

This section examines the per-dataset sensitivity of BoostLLM to two key hyperparameters: the epoch-round allocation and the boosting learning rate.

\Cref{tab:appendix-parameter-sensitivity-rounds} breaks down the round-epoch allocation ablation (discussed in Section~\ref{sec:hyperparam-sensitivity}) by dataset. The granular numbers reveal that the stability observed in the average AP metric holds consistently across individual datasets. For nearly all tasks, an intermediate allocation (e.g., 5 or 6 rounds) provides stable and near-optimal performance. This widespread uniformity confirms that our default setting is robustly applicable across diverse datasets without requiring custom tuning.

Similarly, \Cref{tab:appendix-parameter-sensitivity-boosting-lr} presents the dataset-level results for varying the boosting learning rate ($\eta$). Consistent with the aggregated trend in Section~\ref{sec:hyperparam-sensitivity}, the performance on almost every individual dataset remains nearly flat across a broad range of learning rates. This confirms the main conclusion that the default boosting learning rate is reliably across different data.

\begin{table}[H]
  \caption{Ablation study on the number of boosting rounds (total epochs kept constant).}
  \label{tab:appendix-parameter-sensitivity-rounds}
  \centering
  \resizebox{\textwidth}{!}{%
  \begin{tabular}{lcccccccccccc}
    \toprule
    \textbf{\# rounds} & \textbf{\# epochs} & \textbf{Avg.} & \textbf{bank} & \textbf{blood} & \textbf{calhousing} & \textbf{Car} & \textbf{credit-g} & \textbf{diabetes} & \textbf{heart} & \textbf{income} & \textbf{jungle} \\
    \midrule
    1 & 30 & 74.6 & 64.9 $\pm$ 4.0 & 59.9 $\pm$ 5.6 & 80.3 $\pm$ 2.4 & 91.4 $\pm$ 5.6 & 64.7 $\pm$ 4.1 & 73.3 $\pm$ 2.9 & 84.9 $\pm$ 2.2 & 76.0 $\pm$ 1.7 & 76.1 $\pm$ 3.5 \\
    2 & 15 & 76.9 & 70.0 $\pm$ 2.1 & 61.6 $\pm$ 6.7 & 78.3 $\pm$ 9.9 & 90.3 $\pm$ 6.9 & 68.0 $\pm$ 1.9 & 77.1 $\pm$ 3.5 & 88.2 $\pm$ 3.1 & 79.2 $\pm$ 2.5 & 79.4 $\pm$ 0.6 \\
    3 & 10 & 78.7 & 70.5 $\pm$ 2.0 & 64.7 $\pm$ 6.0 & 82.5 $\pm$ 2.4 & 91.7 $\pm$ 6.6 & 69.9 $\pm$ 2.0 & 77.2 $\pm$ 1.4 & 90.7 $\pm$ 2.8 & 80.1 $\pm$ 1.4 & 80.9 $\pm$ 1.9 \\
    4 & 8 & 79.0 & 71.2 $\pm$ 2.7 & 65.5 $\pm$ 4.8 & 81.7 $\pm$ 2.6 & 90.4 $\pm$ 9.6 & 70.8 $\pm$ 1.1 & 78.7 $\pm$ 2.7 & 91.5 $\pm$ 1.1 & 80.5 $\pm$ 1.5 & 81.0 $\pm$ 2.4 \\
    5 & 6 & 79.7 & 71.3 $\pm$ 2.2 & 65.8 $\pm$ 2.8 & 84.7 $\pm$ 1.6 & 91.4 $\pm$ 5.6 & 71.3 $\pm$ 1.0 & 78.3 $\pm$ 2.8 & 91.3 $\pm$ 2.4 & 82.1 $\pm$ 1.4 & 81.2 $\pm$ 1.6 \\
    6 & 5 & 79.9 & 71.6 $\pm$ 2.6 & 68.3 $\pm$ 5.3 & 84.8 $\pm$ 1.3 & 92.0 $\pm$ 4.6 & 71.1 $\pm$ 1.5 & 77.8 $\pm$ 2.3 & 90.9 $\pm$ 3.5 & 81.2 $\pm$ 1.3 & 81.0 $\pm$ 0.7 \\
    7 & 4 & 79.7 & 71.7 $\pm$ 2.5 & 67.5 $\pm$ 4.3 & 83.9 $\pm$ 2.2 & 90.3 $\pm$ 7.5 & 71.3 $\pm$ 1.3 & 79.1 $\pm$ 3.3 & 91.2 $\pm$ 2.3 & 81.9 $\pm$ 1.6 & 80.8 $\pm$ 1.4 \\
    10 & 3 & 79.7 & 72.1 $\pm$ 1.8 & 66.6 $\pm$ 2.3 & 84.6 $\pm$ 0.9 & 89.2 $\pm$ 4.5 & 71.5 $\pm$ 1.3 & 77.8 $\pm$ 1.1 & 91.0 $\pm$ 2.3 & 82.3 $\pm$ 1.1 & 82.0 $\pm$ 1.8 \\
    15 & 2 & 79.5 & 72.0 $\pm$ 2.9 & 67.8 $\pm$ 4.3 & 84.2 $\pm$ 1.3 & 89.6 $\pm$ 4.3 & 70.3 $\pm$ 1.3 & 78.6 $\pm$ 4.0 & 90.1 $\pm$ 2.8 & 81.6 $\pm$ 1.2 & 81.4 $\pm$ 1.7 \\
    30 & 1 & 77.9 & 70.5 $\pm$ 2.4 & 68.0 $\pm$ 3.3 & 84.2 $\pm$ 1.4 & 81.3 $\pm$ 6.3 & 67.9 $\pm$ 2.6 & 79.3 $\pm$ 3.0 & 89.5 $\pm$ 3.7 & 82.4 $\pm$ 1.0 & 78.1 $\pm$ 2.9 \\
    \bottomrule
  \end{tabular}
  }
\end{table}

\label[appendix]{sec:appendix-parameter-sensitivity-boosting-lr}

\begin{table}[h]
    \caption{Per-dataset average precision for varying boosting learning rates (128-shot, Qwen3-4B).}
    \label{tab:appendix-parameter-sensitivity-boosting-lr}
    \centering
    \resizebox{\textwidth}{!}{%
    \begin{tabular}{ccccccccccc}
        \toprule
        \textbf{lr} & \textbf{Avg.} & \textbf{bank} & \textbf{blood} & \textbf{calhousing} & \textbf{car} & \textbf{credit-g} & \textbf{diabetes} & \textbf{heart} & \textbf{income} & \textbf{jungle} \\
        \midrule
        0.1 & 79.1 & 70.5 ± 3.8 & 66.5 ± 5.1 & 83.9 ± 2.2 & 88.7 ± 7.8 & 70.2 ± 2.4 & 77.9 ± 2.8 & 90.7 ± 2.6 & 81.8 ± 1.2 & 81.5 ± 0.9 \\
        0.3 & 79.7 & 71.3 ± 2.2 & 65.8 ± 2.8 & 84.7 ± 1.6 & 91.4 ± 5.6 & 71.3 ± 1.0 & 78.3 ± 2.8 & 91.3 ± 2.4 & 82.1 ± 1.4 & 81.2 ± 1.6 \\
        0.5 & 79.5 & 71.7 ± 3.2 & 67.2 ± 3.6 & 83.9 ± 2.0 & 89.9 ± 6.1 & 70.9 ± 1.3 & 78.4 ± 3.6 & 90.3 ± 2.3 & 82.0 ± 1.2 & 81.3 ± 1.3 \\
        0.7 & 79.6 & 72.2 ± 2.4 & 67.4 ± 5.2 & 83.1 ± 1.4 & 90.7 ± 6.2 & 70.2 ± 1.3 & 78.2 ± 2.2 & 91.1 ± 2.3 & 81.8 ± 1.2 & 81.6 ± 1.6 \\
        0.9 & 79.1 & 71.6 ± 1.8 & 65.8 ± 4.6 & 84.9 ± 1.3 & 89.4 ± 5.8 & 70.9 ± 1.7 & 77.5 ± 3.7 & 90.0 ± 3.6 & 81.1 ± 1.7 & 80.7 ± 1.2 \\
        1.0 & 78.8 & 71.0 ± 2.5 & 66.4 ± 4.2 & 84.4 ± 0.7 & 88.2 ± 7.2 & 70.0 ± 1.9 & 76.0 ± 3.2 & 90.6 ± 2.6 & 80.9 ± 1.4 & 81.3 ± 1.6 \\
        \bottomrule
    \end{tabular}
    }
\end{table}

\section{Per-round performance of BoostLLM}
\label[appendix]{sec:appendix-grid}

\begin{figure}[H]
    \centering
    \includegraphics[width=\linewidth]{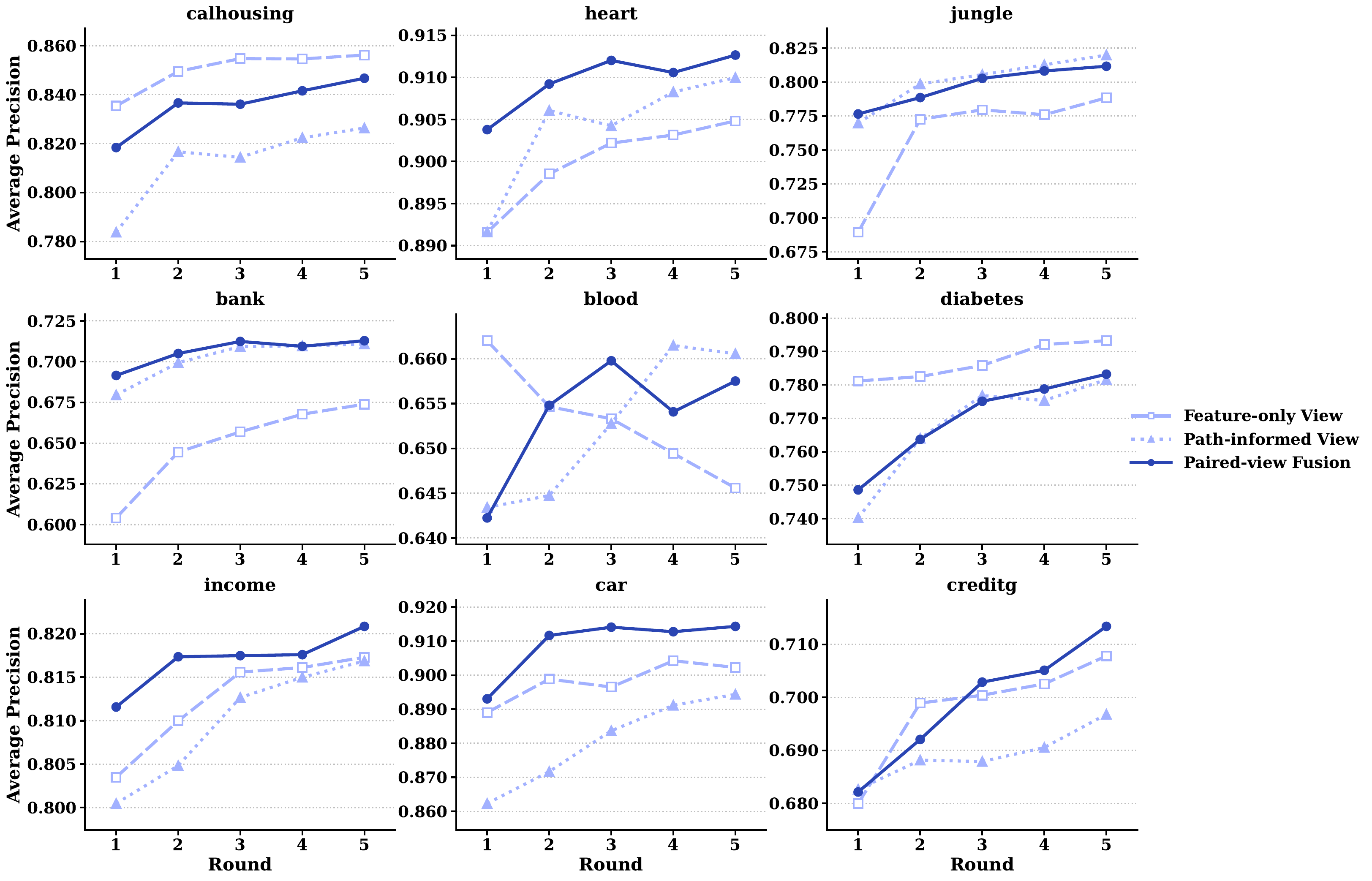}
    \caption{Performance during boosting for each dataset.}
    \label{fig:appendix-3x3-grid}
\end{figure}

Figure~\ref{fig:appendix-3x3-grid} supplements the training dynamics analysis from Section~\ref{sec:boosting-dynamics}. By illustrating the per-round average precision trajectories for all nine datasets individually, we can verify that the monotonic improvement trend seen in the aggregated average faithfully represents the optimization process on each specific task. In nearly all cases, the accumulated model consistently refines its predictions step-by-step. Furthermore, the solid lines (paired-view fusion) invariably track above the dashed lines (path-informed view) and the dot lines (feature-only view) across several datasets, confirming the conclusion that view fusion provides an additive, universally beneficial gain during training.

\section{Per-Dataset View Contribution Ratios}
\label[appendix]{sec:appendix-ratio}

\begin{figure}[H]
  \centering
  \includegraphics[width=\textwidth]{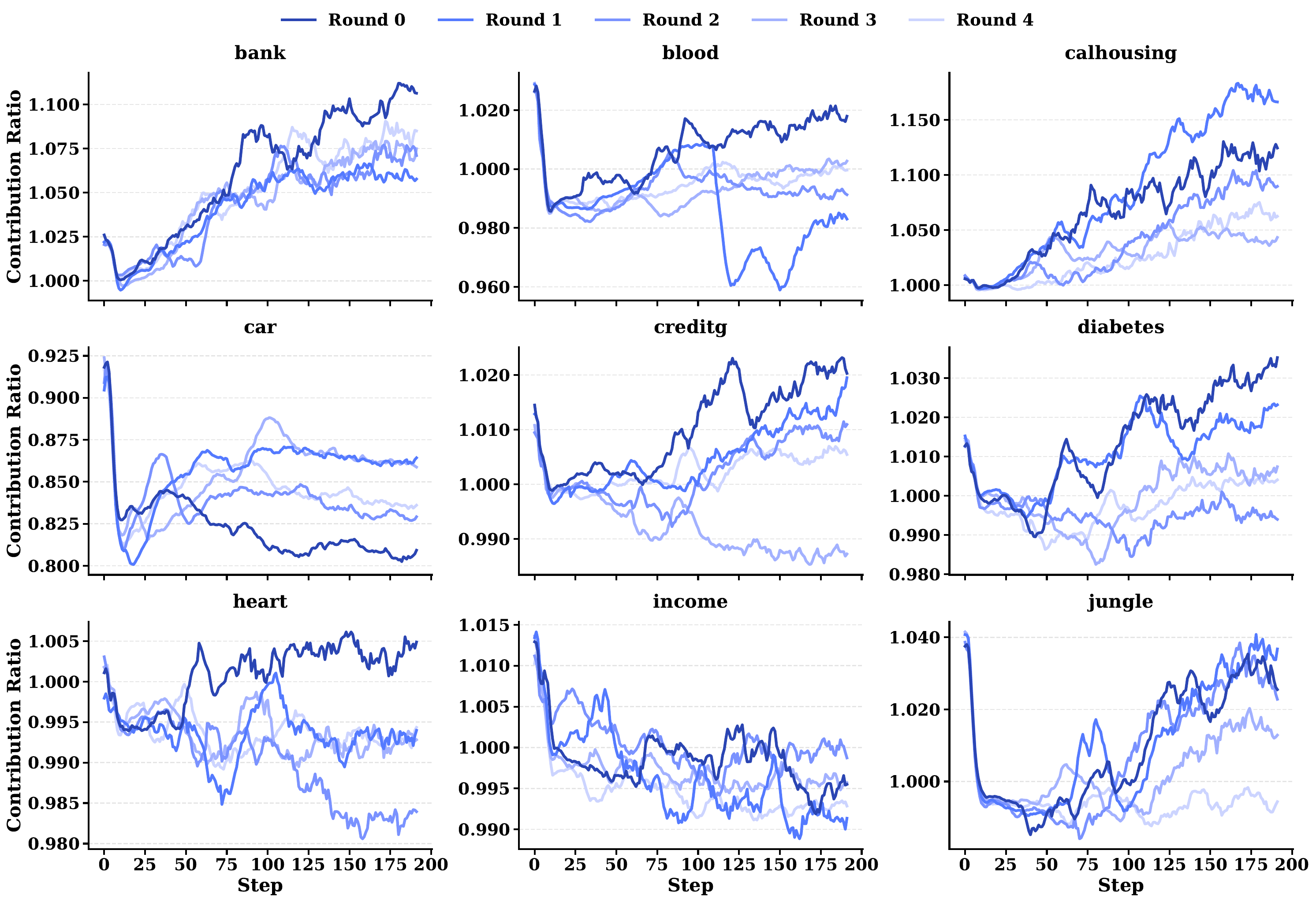}
  \caption{Per-dataset view contribution ratio $\rho$ over training steps (128-shot, Qwen3-4B, 5 rounds). Each subplot shows one dataset; Smaller $\rho$ indicates stronger path-view influence, while larger $\rho$ indicates stronger feature-view influence.}
  \label{fig:appendix-ratio-grid}
\end{figure}

\Cref{fig:appendix-ratio-grid} shows the view contribution ratio $\rho$ for each dataset individually (128-shot, Qwen3-4B, 5 rounds).
The absolute magnitude of $\rho$ varies across datasets---for example, car exhibits $\rho \approx 0.80$--$0.88$, indicating that the path view consistently dominates, while calhousing show $\rho > 1.0$, meaning the feature view carries more weight.
Despite these differences in scale, most datasets display the same qualitative pattern observed in the average (\Cref{fig:magnitude-ratio}): an initial decrease in $\rho$ during early training steps followed by a steady increase, consistent with decision paths acting as early-stage guidance that diminishes as the LLM adapts.

\section{Computational Cost Analysis}
\label[appendix]{sec:appendix-cost}

We discuss the training and inference costs of \model{} relative to the TabLLM baseline.

\paragraph{Training cost.}
As described in \Cref{sec:appendix-hparams}, \model{} allocates the same total 30 epochs as TabLLM, split evenly across $R{=}5$ boosting rounds with 6 epochs per round. Since the backbone, optimizer, and other implementation details are identical, pure boosting (single-view) has equivalent training cost to TabLLM in terms of FLOPs and memory.
The paired-view variant doubles the batch size (the two input views are processed in the same batch) per training step, yielding approximately $2\times$ theoretical training cost. In practice, due to our limited hardware memory, we must apply memory-saving techniques that trade off speed, which makes precise time or time comparisons hardware-dependent.

\paragraph{Inference cost.}
At inference, \model{} requires a sequential forward pass through all $R$ round-specific adapters. With $R{=}5$, the pure boosting variant (no paired view) incurs $5\times$ the inference FLOPs of a single TabLLM forward pass. The full paired-view model processes two prompts per round, leading to ${\approx}10\times$ inference cost.
We note that this is the cost relative to a single TabLLM forward pass of the \emph{same} backbone (Qwen3-4B). In the broader context, training-free baselines such as DeLTa invoke GPT-4o once per sample, whose per-query compute and monetary cost far exceed ten forward passes of a 4B model.
Furthermore, the pure boosting variant---which already surpasses TabLLM by 4.0 AP (\Cref{tab:ablation})---incurs only $5\times$ cost.

\section{Hardware Used}
\label[appendix]{sec:appendix-hardware}

The following hardware configuration was used for all of our experiments. All LLM experiments run on GPUs, while tree-based models run on CPUs .

GPU: NVIDIA L40S (40G); CPU: Intel(R) Xeon(R) Gold 6442Y

\end{document}